\documentclass[10pt,twocolumn,letterpaper]{article}
\ifdefined\pdfminorversion
    \input{glyphtounicode}
\fi
\usepackage[T1]{fontenc}
\usepackage[english]{babel}
\usepackage{times}
\usepackage[margin=0.75in,columnsep=0.28in]{geometry}
\usepackage{amsmath,amssymb,booktabs,graphicx,multirow}
\usepackage{caption}
\usepackage{float}
\usepackage{microtype}
\usepackage[numbers,sort&compress]{natbib}
\usepackage{xcolor}
\definecolor{linkblue}{rgb}{0.12,0.34,0.61}
\usepackage[colorlinks,allcolors=linkblue]{hyperref}
\hypersetup{
    pdftitle={TaskVICL: Task-Conditioned Decoder States for Visual In-Context Learning},
    pdfauthor={Chenyang Wang, Hu Jiamo, Chengrui Wu, Chen Lingyue, Jianguo Chen, Xun Xu, Xulei Yang}
}

\title{TaskVICL: Task-Conditioned Decoder States\\
for Visual In-Context Learning}

\author{%
Chenyang Wang$^{1}$ \quad Hu Jiamo$^{1}$ \quad Chengrui Wu$^{2}$ \quad Chen Lingyue$^{2}$\\
Jianguo Chen$^{3}$ \quad Xun Xu$^{4,5}$ \quad Xulei Yang$^{4}$\\[4pt]
$^{1}$National University of Singapore \quad $^{2}$Nanyang Technological University \quad $^{3}$Sun Yat-sen University\\
$^{4}$Institute of Advanced Intelligence and Computing (IAIC), Singapore\\
$^{5}$Centre for Frontier AI Research (CFAR), A*STAR
}
\date{}
\begin{document}
\maketitle

\begin{abstract}
The usefulness of an in-context demonstration for a vision-language model (VLM) depends on both the query image and the task posed about it. Common retrievers rank examples by image similarity and ignore the task, while task-aware alternatives require answer generation, external encoders or retriever training. We ask whether the frozen target VLM's own decoder states, read before it answers, can select demonstrations directly. TaskVICL represents each image and task by a task-conditioned processing signature: the hidden states at the final input position across all decoder layers. Its training-free variant, TaskVICL-Raw, averages cosine similarities between corresponding layer states and passes the retrieved examples with their reference answers to the same VLM. Across two VLMs and eight classification and visual question answering datasets, Raw achieves the highest average accuracy among training-free retrievers, exceeding the strongest baseline by 3.30 points on Qwen3-VL and 1.41 on InternVL3.5, with gains carrying over to Gemma-3-12B. When the same NICO++ image is asked about its object or its context, Raw retrieves largely different demonstrations for the two questions and outperforms an image retriever given task-matched demonstrations. A lightweight shared projection, TaskVICL-Proj, adds further gains on classification. Controls show that task text at extraction drives most of the gains, and attention analysis describes how it shifts attention toward the classified object. These results support the target VLM's internal states as a practical basis for task-aware demonstration retrieval.
\end{abstract}

\section{Introduction}
\begin{figure}[t]
    \centering
    \includegraphics[width=\linewidth]{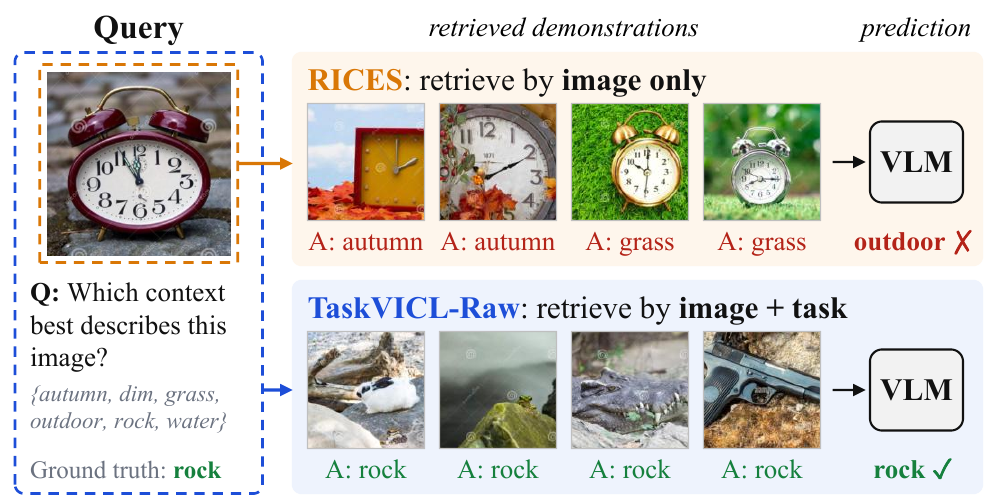}
    \caption{Image-only vs.\ image-and-task retrieval for visual in-context learning. The query is a NICO++ image asked about its context (reference: rock). The dashed boxes mark what each retriever uses. RICES uses the image only (orange), TaskVICL-Raw the image and the task text (blue). The four retrieved demonstrations, each with its question and reference answer, precede the query, and the frozen VLM (Qwen) answers it.}
    \label{fig:visual_icl}

\end{figure}

Visual in-context learning (ICL) lets a vision-language model (VLM) solve a task from a few demonstrations placed in its prompt, without updating its weights \cite{flamingo}. Each demonstration pairs an image and a task with its answer, and the query supplies a new image and task (Fig.~\ref{fig:visual_icl}). Which demonstrations help depends on both parts: the same NICO++ image can be classified by its object or by its context, and the two questions call for different examples (Fig.~\ref{fig:nico_retrieval}).

Existing retrievers only partly address this. RICES ranks examples by image similarity \cite{pica,flamingo} and returns the same demonstrations whatever the question. MMICES adds question-text similarity \cite{mmices}, which reduces to visual similarity when every query shares one instruction, as on single-task classification datasets. Task-aware alternatives pay elsewhere: GPT-MM retrieves with a state taken after answer generation \cite{gptmm}, CIRCLES builds counterfactual contexts at over ten seconds per query in our setup \cite{circles}, and DeTriever trains transformations of decoder states \cite{detriever}. We ask whether the states a target VLM already computes for the query, before it answers, can serve as the retrieval key directly.

At the final input position, the decoder has attended to the whole image and the complete task text, so its hidden states across layers summarize how the model processes this image under this task. We call them a \emph{task-conditioned processing signature} and hypothesize that examples with similar signatures make effective demonstrations. TaskVICL extracts signatures from all decoder layers in one forward pass, without answer generation, external encoders or retriever training. Its training-free Raw variant normalizes each layer state, averages per-layer cosine similarities and retrieves four demonstrations from the full bank with the same rule on every dataset; the same frozen VLM then answers from the selected images, task texts and reference answers. An optional Proj variant learns one shared linear projection from bank labels.

Following recent work \cite{mmices,chen2025provoking,li2025taco,codr,fu2026contextnav,gptmm,circles}, we judge each retriever by the accuracy of the VLM that receives its demonstrations, which also covers open-ended question answering. On Qwen3-VL-4B and InternVL3.5-4B across five fine-grained classification datasets, paired object and context questions on NICO++ and two VQA datasets, Raw has the highest average among training-free retrievers, 81.42 and 68.11, ahead of the strongest baselines by 3.30 and 1.41 points. On NICO++ its neighbors for the two questions about the same image overlap by only 12.08\%, and it beats an image retriever given task-matched demonstrations. Gains carry over to Gemma-3-12B. Our contributions are:
\begin{itemize}
\item TaskVICL, a training-free retriever that compares the frozen target VLM's decoder states at the final input position, layer by layer, before generation.
\item The highest average accuracy among training-free retrievers across two VLMs and eight classification and VQA datasets, with demonstrations that adapt to the question asked about the same image.
\item Controls showing that task text at extraction drives most of the gains, that TaskVICL matches or exceeds retrieval by zero-shot label log-probabilities on all seven classification tasks, and that reference answers correct many initial errors; descriptively, task text also shifts attention toward the classified object.
\end{itemize}

\section{Related Work}
\paragraph{In-context learning with VLMs.}
Flamingo showed that VLMs can solve new tasks from a few interleaved image--text demonstrations \cite{flamingo}, retrieving them by image similarity as proposed in PICa \cite{pica}. This ability is sensitive to the demonstrations: what makes a good visual in-context example is not obvious \cite{zhang2023goodexamples}, and studies of captioning and VQA show that both the choice and the order of demonstrations strongly affect accuracy \cite{yang2023exploringcfg,li2024configurevqa,xu2025introspection}.

\paragraph{Selecting multimodal demonstrations.}
Language ICL retrieves demonstrations by sentence-embedding similarity \cite{kate} or with retrievers trained on language-model feedback \cite{epr,ceil}. Multimodal selectors combine visual and textual cues \cite{mmices}, compose demonstrations from task-oriented image summaries \cite{zhou2024vicL}, or retrieve discriminative examples through counterfactual contexts \cite{circles}. Other methods learn to configure demonstration sequences \cite{leverlm,li2025taco,chen2025provoking}, build compact candidate coresets \cite{codr,chen2025keco} or curate context with an agent \cite{fu2026contextnav}. TaskVICL shares their goal of task-relevant selection but reads relevance from the frozen target VLM itself and keeps the bank's reference answers for prediction.

\paragraph{Model-internal states as retrieval keys.}
GPT-MM retrieves with model predictions \cite{gptmm}; our adaptation uses the final-layer state after answer generation. DeTriever learns layer-specific transformations and aggregation weights from supervised similarity targets for text-to-SQL \cite{detriever}. TaskVICL instead compares states at corresponding depths before generation and needs no supervision in its Raw form. More broadly, VLM2Vec learns instruction-conditioned multimodal embeddings \cite{vlm2vec}, E5-V uses the final-token state of a prompted multimodal LLM as a universal embedding \cite{e5v}, and multimodal task vectors compress demonstrations into attention-head activations \cite{mtv}. These works use internal states as general-purpose embeddings or as substitutes for demonstrations, whereas TaskVICL uses the prediction model's own states to select labeled demonstrations.

\paragraph{Interpreting visual processing.}
Token Activation Maps attribute generated tokens to multimodal inputs \cite{tam}, while work on attention and saliency cautions against reading internal measurements as explanations of decisions \cite{jainwallace,adebayo}. We therefore use attention only to describe how processing changes with the task across depth, and measure demonstration utility separately through final predictions.

\section{Method}
\label{sec:method}
\subsection{Task-conditioned processing signature}
Let $F$ be a frozen VLM and $\mathcal B=\{(x_i,t_i,y_i)\}_{i=1}^{N}$ a labeled bank, where $x_i$ is an image, $t_i$ its task text and $y_i$ its reference answer. A query supplies $(x_q,t_q)$. For classification, the task text contains the instruction and the label list. For visual question answering (VQA), it contains the question and any answer options.

For the native input sequence $u(x,t)$, let $T(x,t)$ be the final input position, including the assistant prefix. We extract
\begin{equation}
h_\ell(x,t)=F_\ell\bigl(u(x,t)\bigr)_{T(x,t)},\quad \ell=1,\ldots,L,
\label{eq:states}
\end{equation}
from all decoder layers in a single forward pass before generation, excluding the embedding output and keeping native image processing. Qwen gives $L=36$ states of width $d=2560$.

\begin{figure*}[!t]
\centering
\includegraphics[width=\textwidth]{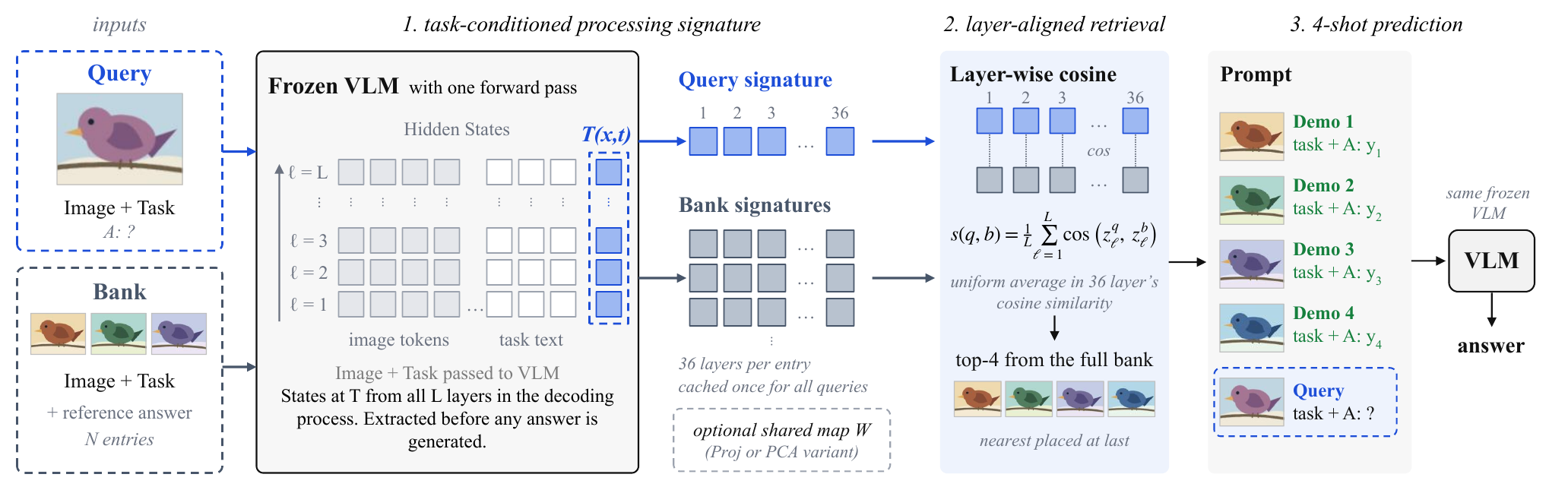}
\caption{TaskVICL retrieves demonstrations with states of the frozen prediction VLM. Each decoder layer contributes its state at the final input position. Raw averages per-layer cosine similarities. Proj adds a shared projection learned from bank labels, and PCA gives an unsupervised projection for question answering. The selected examples enter the final four-shot prompt.}
\label{fig:pipeline}

\end{figure*}

Under causal attention, the final input position follows the complete image and task text and can draw on both. Its states therefore change with the requested prediction even when the image stays the same. On single-task datasets all entries share one task text, which fixes the instruction and label list. NICO++ and question answering vary the task across entries. We call $\{h_\ell(x,t)\}_{\ell=1}^{L}$ a \textbf{task-conditioned processing signature}. The final input position also prepares the first answer token, so the signature reflects how the model is about to respond to this image and task. Reading it requires no answer generation, and the bank's reference answers still guide the final prediction. Figure~\ref{fig:pipeline} summarizes the pipeline.

\subsection{Training-free retrieval}
For an input $a=(x,t)$, each state is normalized separately as $z_\ell(a)=h_\ell(a)/\|h_\ell(a)\|_2$. TaskVICL-Raw compares corresponding layers with
\begin{equation}
s_{\mathrm{Raw}}(a,b)=\frac{1}{L}\sum_{\ell=1}^{L}z_\ell(a)^\top z_\ell(b).
\label{eq:score}
\end{equation}
Every layer contributes equally, and the score equals the cosine similarity between concatenated normalized states.

We cache the bank signatures, select the four highest-scoring entries and order them from lower to higher similarity, so the nearest example comes last. Each demonstration contributes its image, original task text and reference answer, followed by the query, and the same frozen VLM predicts the answer. In every ablation that alters the extraction text, the final prompt keeps the original task.

\subsection{Shared projection}
TaskVICL-Proj keeps the signature and the uniform layer average and applies one matrix $W\in\mathbb R^{p\times d}$ with $p=256$ to every layer. With $\nu(v)=v/\|v\|_2$, the score is
\begin{equation}
s_W(a,b)=\frac{1}{L}\sum_{\ell=1}^{L}\nu(Wz_\ell(a))^\top\nu(Wz_\ell(b)).
\label{eq:projection}
\end{equation}
For Qwen, $W$ has 655{,}360 parameters and reduces the bank cache tenfold. The VLM stays frozen.

For classification, $W$ is trained so that bank examples with the same label score higher. For each anchor $a$ in a balanced batch $\mathcal A$, let $P(a)$ contain the other examples sharing its label. We minimize a supervised contrastive loss \cite{supcon},
\begin{equation}
\mathcal L(W)=-\frac{1}{|\mathcal A|}\sum_{a\in\mathcal A}
\log\frac{\sum_{b\in P(a)}\exp(s_W(a,b)/\tau)}
{\sum_{c\in\mathcal A\setminus\{a\}}\exp(s_W(a,c)/\tau)}.
\end{equation}
Single-task datasets use class labels, and NICO++ uses the 66 groups formed by task and native label. Training runs 1500 steps with learning rate $3\times10^{-4}$, $\tau=0.07$ and batches of 16 label groups with four examples each. Ten bank images per class, or per object class with both questions on NICO++, are held out, and we keep the earliest checkpoint with the best top-four label agreement on them.

Question answering has no fixed label set, and few training answers repeat often enough to form groups. OK-VQA has only five answers shared by at least 14 bank questions. For ScienceQA and OK-VQA we therefore replace the learned map with a 256-dimensional PCA basis fitted on bank states without using reference answers, and keep the normalization and layer average of Eq.~\eqref{eq:projection}. We report this unsupervised variant as TaskVICL-PCA in Table~\ref{tab:main}, with a supervised OK-VQA map in Appendix~\ref{sec:supp_projection}.

\begin{table*}[t]
\centering
\caption{Prediction results with 4 shots, in percent. Classification and ScienceQA report accuracy and OK-VQA the official consensus score. Avg.\ weights each dataset equally after averaging the two NICO++ tasks. \emph{Training-free} retrievers use no parameters fitted on the bank. \emph{Training-based} retrievers fit parameters on the bank, with labels for DeTriever and Proj and without labels for TaskVICL-PCA on question answering. Bold marks the best result within each segment. GPT-MM uses only its retrieval component, and MMICES uses the target VLM\textquotesingle s visual features. InternVL zero-shot ScienceQA counts 31.9\% invalid outputs as errors.}
\label{tab:main}
\small
\setlength{\tabcolsep}{3pt}
\renewcommand{\arraystretch}{0.85}
\setlength{\aboverulesep}{0.25ex}
\setlength{\belowrulesep}{0.25ex}
\begin{tabular*}{\textwidth}{@{\extracolsep{\fill}}c l rrrrrrr rr r@{}}
\toprule
 & & \multicolumn{7}{c}{Classification} & \multicolumn{2}{c}{Question answering} & Overall \\
\cmidrule(lr){3-9} \cmidrule(lr){10-11} \cmidrule(l){12-12}
 & & \multicolumn{2}{c}{NICO++} & & & & & & & & \\
\cmidrule(lr){3-4}
VLM & Method & Object & Context & DTD & Aircraft & CUB & Dogs & Pets & ScienceQA & OK-VQA & Avg. \\
 & Query count & 1080 & 1080 & 1880 & 3333 & 5794 & 8580 & 3669 & 2017 & 5046 & -- \\
\midrule
\multirow{14}{*}{\rotatebox[origin=c]{90}{\textbf{Qwen3-VL-4B}}}
  & \multicolumn{11}{l}{\textit{Training-free}} \\
\cmidrule(l){2-12}
  & Zero-shot & 91.57 & 38.89 & 67.07 & 32.31 & 37.18 & 54.39 & 82.47 & 90.73 & 58.74 & 61.02 \\
  & Random-4 & 90.74 & 39.17 & 58.78 & 33.21 & 28.34 & 49.69 & 70.18 & 89.84 & 57.14 & 56.52 \\
  & RICES (CLIP ViT-B/32)~\cite{pica,clip} & 92.78 & 45.28 & 78.72 & 59.68 & 71.21 & 74.32 & 91.52 & 89.99 & 57.63 & 74.01 \\
  & RICES (CLIP ViT-L/14)~\cite{pica,clip} & 93.06 & 46.39 & 80.80 & 68.47 & 80.93 & 80.35 & 93.30 & 89.99 & 57.17 & 77.59 \\
  & RICES (DINOv3)~\cite{pica,dinov3} & 91.94 & 46.39 & 82.50 & 57.76 & 65.36 & 76.59 & 92.20 & 90.28 & 57.36 & 73.90 \\
  & GPT-MM~\cite{gptmm} & 93.70 & 61.67 & 81.60 & 69.37 & 72.64 & 78.68 & 92.86 & 92.51 & 59.64 & 78.12 \\
  & CIRCLES~\cite{circles} & 93.24 & 52.59 & 80.85 & 67.57 & \textbf{81.91} & 79.36 & 93.32 & 90.28 & 52.39 & 77.32 \\
  & MMICES~\cite{mmices} & 94.07 & 61.57 & 82.61 & 65.35 & 78.34 & 77.16 & 91.63 & 90.28 & 56.97 & 77.52 \\
  & \textbf{TaskVICL-Raw} (Ours) & \textbf{94.54} & \textbf{63.70} & \textbf{84.04} & \textbf{73.84} & 81.50 & \textbf{82.97} & \textbf{94.96} & \textbf{94.20} & \textbf{60.75} & \textbf{81.42} \\
\cmidrule(l){2-12}
  & \multicolumn{11}{l}{\textit{Training-based}} \\
\cmidrule(l){2-12}
  & DeTriever~\cite{detriever} & 92.69 & 59.63 & 78.24 & 68.38 & 68.81 & 77.68 & 92.40 & 93.36 & 58.92 & 76.74 \\
  & \textbf{TaskVICL-Proj} (Ours) & \textbf{94.44} & \textbf{65.28} & \textbf{85.90} & \textbf{78.67} & \textbf{83.36} & \textbf{85.20} & \textbf{95.42} & -- & -- & -- \\
  & \textbf{TaskVICL-PCA} (Ours) & -- & -- & -- & -- & -- & -- & -- & \textbf{94.35} & \textbf{60.79} & -- \\
\midrule
\multirow{14}{*}{\rotatebox[origin=c]{90}{\textbf{InternVL3.5-4B}}}
  & \multicolumn{11}{l}{\textit{Training-free}} \\
\cmidrule(l){2-12}
  & Zero-shot & 89.72 & 35.37 & 53.14 & 14.04 & 14.93 & 18.55 & 50.15 & 65.20 & 32.14 & 38.84 \\
  & Random-4 & 87.69 & 41.57 & 34.95 & 9.63 & 10.41 & 13.61 & 33.69 & 93.36 & 47.62 & 38.49 \\
  & RICES (CLIP ViT-B/32)~\cite{pica,clip} & 89.54 & 47.69 & 73.24 & 43.02 & 56.80 & 55.21 & 78.69 & 92.61 & 47.62 & 64.48 \\
  & RICES (CLIP ViT-L/14)~\cite{pica,clip} & 90.46 & 50.83 & 73.62 & \textbf{46.41} & 55.26 & 61.86 & 81.25 & 93.31 & 47.72 & 66.26 \\
  & RICES (DINOv3)~\cite{pica,dinov3} & 90.65 & 50.37 & \textbf{76.60} & 43.86 & 54.06 & 62.66 & 82.83 & 92.86 & 48.24 & 66.45 \\
  & GPT-MM~\cite{gptmm} & 90.09 & 62.04 & 70.05 & 37.56 & 46.25 & 48.88 & 75.39 & 92.81 & 50.55 & 62.20 \\
  & CIRCLES~\cite{circles} & 91.39 & 55.19 & 76.12 & 43.53 & 54.95 & 62.53 & 82.17 & 92.27 & 48.77 & 66.70 \\
  & MMICES~\cite{mmices} & 89.91 & 59.63 & 75.96 & 33.45 & 32.03 & 40.65 & 60.48 & 92.81 & 48.47 & 57.33 \\
  & \textbf{TaskVICL-Raw} (Ours) & \textbf{91.48} & \textbf{62.31} & 75.96 & 42.33 & \textbf{57.58} & \textbf{63.04} & \textbf{84.33} & \textbf{93.41} & \textbf{51.31} & \textbf{68.11} \\
\cmidrule(l){2-12}
  & \multicolumn{11}{l}{\textit{Training-based}} \\
\cmidrule(l){2-12}
  & DeTriever~\cite{detriever} & 89.63 & 60.65 & 72.55 & 40.71 & 54.45 & 62.09 & 85.99 & 92.27 & 49.95 & 66.64 \\
  & \textbf{TaskVICL-Proj} (Ours) & \textbf{91.94} & \textbf{63.52} & \textbf{77.50} & \textbf{49.14} & \textbf{67.45} & \textbf{67.84} & \textbf{88.61} & -- & -- & -- \\
  & \textbf{TaskVICL-PCA} (Ours) & -- & -- & -- & -- & -- & -- & -- & \textbf{93.26} & \textbf{51.61} & -- \\
\bottomrule
\end{tabular*}

\end{table*}

\section{Experiments}
\label{sec:setup}
\subsection{Setup}
We evaluate Qwen3-VL-4B-Instruct \cite{qwen3vl} and InternVL3.5-4B \cite{internvl35} on eight datasets. NICO++~\cite{nico} provides paired object and context classification on the same images. We draw a seeded sample of its official training images, stratified over 60 objects and six contexts, with 3960 bank images and 1080 disjoint query images. Every image carries both questions, giving 7920 bank entries and 2160 queries in one mixed bank. DTD \cite{dtd}, FGVC Aircraft \cite{aircraft}, CUB \cite{cub}, Stanford Dogs \cite{dogs} and Oxford Pets \cite{pets} cover single-task classification, and the 2{,}017 image questions of ScienceQA \cite{scienceqa} and all 5046 OK-VQA validation questions \cite{okvqa} cover question answering. Classification and ScienceQA report accuracy, and OK-VQA uses the official ten-answer consensus score.

Every ICL method retrieves four demonstrations from the same bank. Zero-shot uses no demonstrations, and Random-4 samples them uniformly. All classification predictions, including zero-shot, use legal-label decoding over the listed categories. RICES ranks bank images by cosine similarity of frozen image features \cite{pica,flamingo}, computed with CLIP ViT-B/32 or ViT-L/14 \cite{clip} or with DINOv3 ViT-B/16 \cite{dinov3}. RICES without a qualifier denotes the CLIP ViT-B/32 variant. GPT-MM retrieves with the target VLM's final-layer state after it generates an answer \cite{gptmm}. Our DeTriever adaptation learns per-layer nonlinear maps and layer weights on the target VLM's states, using bank input and reference-answer representations as similarity targets \cite{detriever}. CIRCLES adds retrieval guided by counterfactual attributes generated with the target VLM \cite{circles}. MMICES preselects 32 bank images with the target VLM's own visual features and keeps the four whose question text is most similar under CLIP ViT-B/32 \cite{mmices}. Appendix~\ref{sec:supp_baselines} details all baselines.

Paired 95\% confidence intervals use 2000 query-bootstrap resamples \cite{efron}, and NICO++ resamples images with both questions kept together. Invalid outputs remain in every denominator, and differences are computed from unrounded scores.

\subsection{Main results}
\label{sec:results}
Table~\ref{tab:main} evaluates each retriever through the final VLM prediction. Among training-free retrievers, Raw has the highest average on both VLMs, 81.42 on Qwen and 68.11 on InternVL. The strongest training-free baselines, GPT-MM on Qwen and CIRCLES on InternVL, trail it by 3.30 and 1.41 points. CLIP ViT-L/14 RICES trails it by 3.83 points on Qwen with a 95\% interval of $[3.38,4.27]$ and by 1.85 points on InternVL with interval $[1.41,2.26]$. With task-matched RICES on NICO++, these two gaps become 2.84 and 1.25 points. On NICO++, Raw has the best mean over the two questions on both VLMs. Its gains come mainly from context questions, where it reaches 63.70 on Qwen against at most 46.39 for the RICES variants. Image retrievers return the same images for both questions, while Raw's neighbors share the query's question type, which Section~\ref{sec:nico_paired} controls for with task-matched RICES. MMICES also matches the question type through its text reranking. On single-task datasets this text is constant, so MMICES ranks by the VLM's visual features alone. On InternVL these come from one untiled view, and MMICES falls well below Raw.

On Qwen, Raw exceeds every baseline in eight of the nine result columns, and only CIRCLES is higher on CUB (81.91 vs.\ 81.50). On InternVL, Raw again has the highest average among training-free retrievers. Raw is numerically highest in seven of nine columns among training-free methods. On question answering, all three RICES variants fall below zero-shot on Qwen, while Raw improves to 94.20 on ScienceQA and 60.75 on OK-VQA. On InternVL ScienceQA, all ICL methods lie within 1.2 points, and Raw's gain over RICES has an interval including zero.

Proj improves 13 of 14 classification scores over Raw and lowers only Qwen NICO++ object accuracy. On question answering, PCA changes Raw by at most 0.30 points with intervals including zero, while reducing storage tenfold (Appendix~\ref{sec:supp_projection}). Results over three runs, one to eight demonstrations and the label precision of retrieved examples appear in Appendix~\ref{sec:supp_additional}.

\subsection{Task changes on the same image}
\label{sec:nico_paired}
\begin{figure*}[t]
\centering
\includegraphics[width=\textwidth]{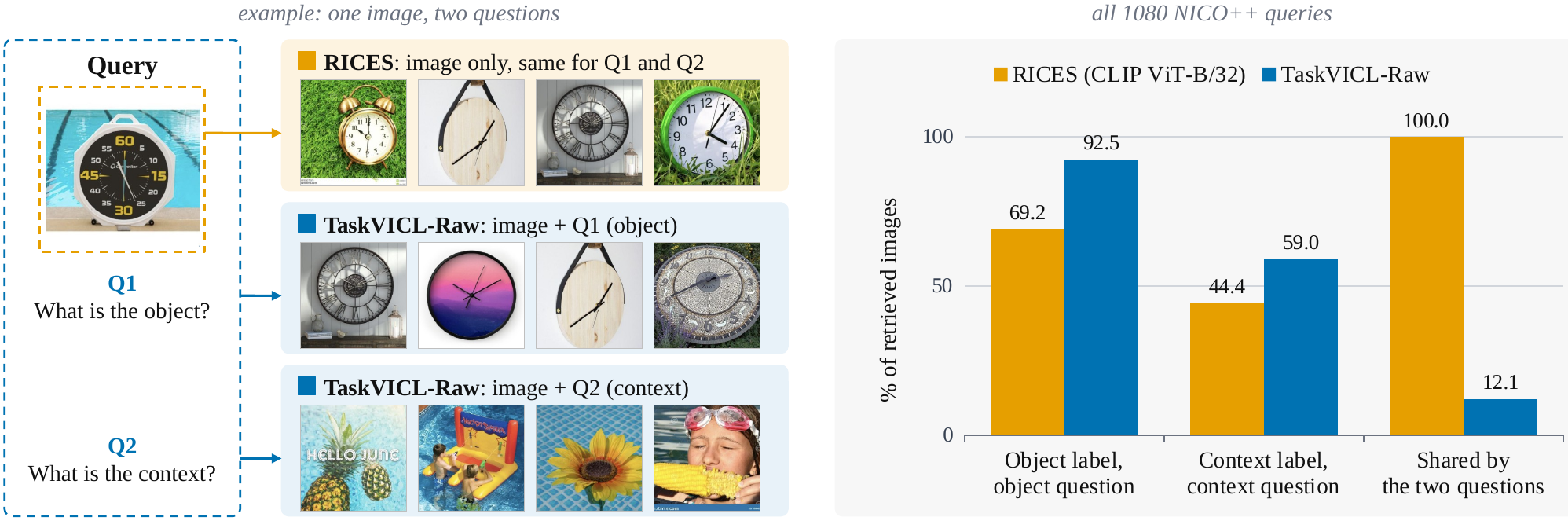}
\caption{Retrieval on paired NICO++ tasks with Qwen. Left, an example query. RICES retrieves with the image alone and selects the same images for both questions. TaskVICL-Raw retrieves with the image and the question together and selects different demonstrations for each. Right, over all 1080 queries, the share of retrieved images whose native label matches the query for the asked question, and the share retrieved for both questions.}
\label{fig:nico_retrieval}

\end{figure*}

Figure~\ref{fig:nico_retrieval} shows how retrieval responds when the same image is asked a different question. RICES returns the same images for both questions. Raw returns different images, and only 12.08\% of its neighbors are shared by the two questions. The native labels of the retrieved images follow the question. Under object questions, 92.50\% of Raw's neighbors share the query object, against 69.19\% for RICES. Under context questions, 58.98\% share the query context, against 44.42\%.

Part of this gain could come from retrieving demonstrations of the same question type. A task-matched RICES control keeps the RICES images and their order and supplies each with the question and bank answer of the query task, for all three RICES encoders. Task matching helps every encoder on both VLMs (Table~\ref{tab:nico}), and TaskVICL-Raw still exceeds the strongest task-matched RICES control, CLIP ViT-L/14, by 1.48 points on both VLMs, with paired 95\% intervals of [0.19, 2.73] on Qwen and [0.19, 2.78] on InternVL. Task matching closes 7.17 of the 10.09-point gap between CLIP ViT-B/32 RICES and Raw on Qwen. The remaining gap reflects which images Raw retrieves for each question. 

\begin{table}[t]
\centering\small
\caption{NICO++ accuracy, mean of object and context questions. \emph{Image} uses RICES neighbors as retrieved; \emph{+ Task} keeps the same images and order but pairs each with the query's question and its bank answer. Image rows for ViT-L/14 and DINOv3 are averaged from Table~\ref{tab:main}.}
\label{tab:nico}
\setlength{\tabcolsep}{2pt}
\renewcommand{\arraystretch}{0.85}
\setlength{\aboverulesep}{0.2ex}
\setlength{\belowrulesep}{0.2ex}
\begin{tabular*}{\linewidth}{@{\extracolsep{\fill}}lrrrr@{}}
\toprule
 & \multicolumn{2}{c}{Qwen} & \multicolumn{2}{c}{InternVL}\\
\cmidrule(lr){2-3} \cmidrule(l){4-5}
Method & Image & + Task & Image & + Task\\
\midrule
RICES (CLIP ViT-B/32)~\cite{pica,clip} & 69.03 & 76.20 & 68.61 & 73.75\\
RICES (CLIP ViT-L/14)~\cite{pica,clip} & 69.73 & 77.64 & 70.65 & 75.42\\
RICES (DINOv3)~\cite{pica,dinov3} & 69.17 & 76.76 & 70.51 & 75.37\\
\midrule
GPT-MM~\cite{gptmm} & & 77.69 & & 76.06\\
MMICES~\cite{mmices} & & 77.82 & & 74.77\\
\textbf{TaskVICL-Raw} (Ours) & & \textbf{79.12} & & \textbf{76.90}\\
\bottomrule
\end{tabular*}

\end{table}

\subsection{Model scale and cost}
\label{sec:scale}
\paragraph{Larger model.}
We run Gemma-3-12B-IT \cite{gemma3}, a larger model from another family, on NICO++, DTD and OK-VQA with the same protocol. Raw exceeds RICES by 5.79 points on NICO++, mostly through a 10.00-point gain on context questions, and by 3.65 on OK-VQA (Table~\ref{tab:gemma}). On DTD it is 1.12 points higher with an interval including zero. Appendix~\ref{sec:supp_qwen8b} reports results on the same three datasets for Qwen3-VL-8B-Instruct.

\begin{table}[t]
\centering\small
\caption{Gemma-3-12B-IT with 4 shots. Accuracy, except the OK-VQA consensus score. RICES~\cite{pica} uses CLIP ViT-B/32~\cite{clip}, and Raw is TaskVICL-Raw (Ours). $\Delta$ is Raw $-$ RICES with paired 95\% intervals; gray marks an interval that includes zero.}
\label{tab:gemma}
\setlength{\tabcolsep}{3.5pt}
\renewcommand{\arraystretch}{0.85}
\setlength{\aboverulesep}{0.2ex}\setlength{\belowrulesep}{0.2ex}
\begin{tabular*}{\linewidth}{@{\extracolsep{\fill}}lrrrl@{}}
\toprule
Dataset & Zero-shot & RICES & Raw & \multicolumn{1}{c}{$\Delta$}\\
\midrule
NICO++ object & 90.83 & 91.11 & \textbf{92.69} & $+1.57$ {\scriptsize$[0.37, 2.78]$}\\
NICO++ context & 37.41 & 54.44 & \textbf{64.44} & $+10.00$ {\scriptsize$[7.13, 12.69]$}\\
\quad \textit{mean} & 64.12 & 72.78 & \textbf{78.56} & $+5.79$ {\scriptsize$[4.30, 7.36]$}\\
\midrule
DTD & 67.87 & 81.54 & \textbf{82.66} & \textcolor{gray}{$+1.12$ {\scriptsize$[-0.43, 2.66]$}}\\
OK-VQA & 34.73 & 54.28 & \textbf{57.92} & $+3.65$ {\scriptsize$[2.37, 4.84]$}\\
\bottomrule
\end{tabular*}

\end{table}

\paragraph{Cost.}
Bank states are extracted once and cached, and each query needs one extra forward pass before generation. On an RTX 4090, Raw retrieval takes 267\,ms per query including preprocessing, against 675\,ms for four-shot generation and over ten seconds for CIRCLES. Proj reduces the CUB bank cache from 2107 to 211\,MiB (Appendix~\ref{sec:supp_cost}).

\section{Ablations and Analysis}
\label{sec:ablation}
This section isolates the sources of TaskVICL's gains by varying what is extracted, where it is read from, and which answers accompany the retrieved demonstrations. All analyses use Qwen unless stated otherwise.

\subsection{Task information during extraction}
We vary the text supplied during state extraction while the final prompt stays fixed. Task uses the original instruction or question. Generic replaces it with \emph{Describe the image briefly}, and Generic + List appends the full class list (examples in Appendix~\ref{sec:supp_source}). Table~\ref{tab:repr_merged} shows that the class list alone raises Generic accuracy by 5.21 to 22.89 points, and the full task text is best on DTD and CUB. On OK-VQA, task text raises Raw from 54.07 to 60.75 (Sec.~\ref{sec:answers}), so its benefit extends beyond the class list.

\begin{table}[t]
\centering\small
\caption{Retrieval keys on Qwen. Each group varies one part of TaskVICL-Raw (all layers, final input position, task prompt). Generic uses \emph{Describe the image briefly}; + List appends the class list. GPT-MM uses the final-layer state after generation.}
\label{tab:repr_merged}
\setlength{\tabcolsep}{4pt}
\renewcommand{\arraystretch}{0.85}
\setlength{\aboverulesep}{0.2ex}
\setlength{\belowrulesep}{0.2ex}
\begin{tabular*}{\linewidth}{@{\extracolsep{\fill}}lrrr@{}}
\toprule
Retrieval key & DTD & CUB & OK-VQA\\
\midrule
\textbf{TaskVICL-Raw} (Ours) & \textbf{84.04} & \textbf{81.50} & 60.75\\
\midrule
\multicolumn{4}{@{}l}{\textit{Extraction text}}\\
\cmidrule{1-4}
\quad Generic & 78.56 & 46.29 & 54.07\\
\quad Generic + List & 83.78 & 69.18 & --\\
\midrule
\multicolumn{4}{@{}l}{\textit{Decoder depth}}\\
\cmidrule{1-4}
\quad Early & 71.01 & 32.05 & 58.07\\
\quad Middle & 83.24 & 78.05 & 59.22\\
\quad Late & 83.35 & 81.00 & \textbf{60.77}\\
\quad Last & 81.72 & 79.93 & 59.05\\
\midrule
\multicolumn{4}{@{}l}{\textit{Other representations}}\\
\cmidrule{1-4}
\quad GPT-MM~\cite{gptmm} & 81.60 & 72.64 & 59.64\\
\quad RICES~\cite{pica} (Qwen vision tower) & 82.66 & 78.51 & 57.39\\
\quad VLM2Vec~\cite{vlm2vec} & 81.65 & 70.26 & 60.69\\
\quad Caption BGE-M3~\cite{bge} & 78.35 & 55.68 & 57.38\\
\quad Caption BM25~\cite{bm25} & 78.88 & 45.31 & 57.04\\
\bottomrule
\end{tabular*}

\end{table}

\subsection{Decoder depth}
\label{sec:layers}
We keep the final input position and vary the layers used. Last is the final decoder layer, Early, Middle and Late are consecutive groups of 12 of Qwen's 36 layers, and All is Raw.

All exceeds Last by 2.32 points on DTD and 1.57 on CUB, with intervals of $[1.60,3.04]$ and $[1.01,2.13]$, and matches Late on OK-VQA with a difference of $-0.02$ and interval $[-0.66,0.58]$. Uniform averaging over all layers needs no layer selection and matches or exceeds every layer group within the intervals.

\subsection{Zero-shot label beliefs}
\label{sec:belief}
The final input position prepares the first answer token, so Raw might only retrieve examples that the model would label in the same way. We test this with a \textit{belief retriever} that uses the same input and scores the log-likelihood of every legal label without demonstrations. Prob.\ compares softmax label probabilities by cosine similarity, and Log.\ compares mean-centered log-probabilities. On NICO++, belief retrieval searches only bank entries of the query's question type. The top belief label equals the zero-shot prediction for 94.8\% to 99.1\% of images, so these vectors encode the model's zero-shot answer.

\begin{table}[t]
\centering\small
\caption{TaskVICL-Raw (Ours) against zero-shot label beliefs on Qwen. Vote is the majority label of Raw's neighbors. $\Delta$ is Raw $-$ Log.\ with paired 95\% intervals; gray marks an interval that includes zero.}
\label{tab:belief}
\setlength{\tabcolsep}{1.5pt}
\renewcommand{\arraystretch}{0.85}
\setlength{\aboverulesep}{0.2ex}\setlength{\belowrulesep}{0.2ex}
\begin{tabular*}{\linewidth}{@{\extracolsep{\fill}}lrrrrl@{}}
\toprule
Dataset & Raw & Vote & Prob. & Log. & \multicolumn{1}{c}{$\Delta$}\\
\midrule
DTD & \textbf{84.04} & 82.13 & 76.01 & 81.70 & $+2.34$ {\scriptsize$[1.17, 3.62]$}\\
Aircraft & \textbf{73.84} & 70.33 & 59.59 & \textbf{73.84} & \textcolor{gray}{$0.00$ {\scriptsize$[-1.29, 1.32]$}}\\
CUB & \textbf{81.50} & 78.41 & 64.20 & 80.88 & \textcolor{gray}{$+0.62$ {\scriptsize$[-0.22, 1.47]$}}\\
Dogs & \textbf{82.97} & 81.75 & 73.41 & 81.77 & $+1.20$ {\scriptsize$[0.56, 1.86]$}\\
Pets & \textbf{94.96} & 93.10 & 91.99 & 94.49 & \textcolor{gray}{$+0.46$ {\scriptsize$[-0.08, 0.98]$}}\\
NICO++ obj. & \textbf{94.54} & 93.15 & 93.52 & 92.50 & $+2.04$ {\scriptsize$[1.11, 2.97]$}\\
NICO++ ctx. & \textbf{63.70} & 62.59 & 51.67 & 53.80 & $+9.91$ {\scriptsize$[7.69, 12.41]$}\\
\bottomrule
\end{tabular*}

\end{table}

Raw exceeds Prob.\ on all seven tasks by 1.02 to 17.29 points, with every interval excluding zero (Table~\ref{tab:belief}). The gap comes from queries that the model misclassifies without demonstrations, where Raw leads Prob.\ by 16.17 to 24.67 points on the single-task datasets, while the two differ by at most 4.83 points on initially correct queries. Log.\ matches Raw on Aircraft, CUB and Pets, yet Raw still leads on DTD, Dogs and both NICO++ questions, most clearly by 9.91 points on context questions with only six labels. The decoder states therefore carry information beyond the zero-shot label distribution (Appendix~\ref{sec:supp_belief}). Where accuracy matches, the retrieved examples still differ, and Log.\ shares at most 50.8\% of its neighbors with Raw on any task. Belief retrieval also needs a fixed label list and one likelihood per label, so it cannot serve open-ended questions such as OK-VQA, where Raw applies unchanged. Baldassini et al.\ find that ICL with image retrieval often fails to beat label voting over the retrieved examples \cite{baldassini2024}. Majority voting over Raw's neighbors without the VLM reaches 81.14 on the five single-task datasets, and ICL adds 2.32 points and exceeds voting on all seven tasks (Appendix~\ref{sec:supp_voting}).

\subsection{External representations}
We next compare Raw with representations outside the target decoder (Table~\ref{tab:repr_merged}). VLM2Vec is a trained instruction-conditioned multimodal embedder that receives the same image and task text \cite{vlm2vec}. The caption selectors join an offline generic caption from Qwen3-VL-4B with the task text and rank it with BGE-M3 embeddings \cite{bge} or BM25 \cite{bm25}.

Raw exceeds VLM2Vec by 2.39 points on DTD and 11.24 on CUB, with intervals of $[1.01,3.83]$ and $[10.10,12.39]$, and the two are on par on OK-VQA with interval $[-0.83,0.97]$. Raw exceeds both caption selectors on all three datasets, and all six paired intervals exclude zero. Qwen's vision tower trails Raw by 1.38 points on DTD, 2.99 on CUB and 3.36 on OK-VQA, which suggests that the decoder's joint processing of image and task contributes to the gain (paired differences in Appendix~\ref{sec:supp_layers}). These controls separate Raw from post-generation and general-purpose embeddings. Pre-generation states of all layers exceed the GPT-MM state, decoder states exceed the vision tower, and task text exceeds a generic prompt (Table~\ref{tab:repr_merged}).
\subsection{Bank answers and corrective context}
\label{sec:answers}
\begin{figure*}[t]
\centering
\includegraphics[width=\textwidth]{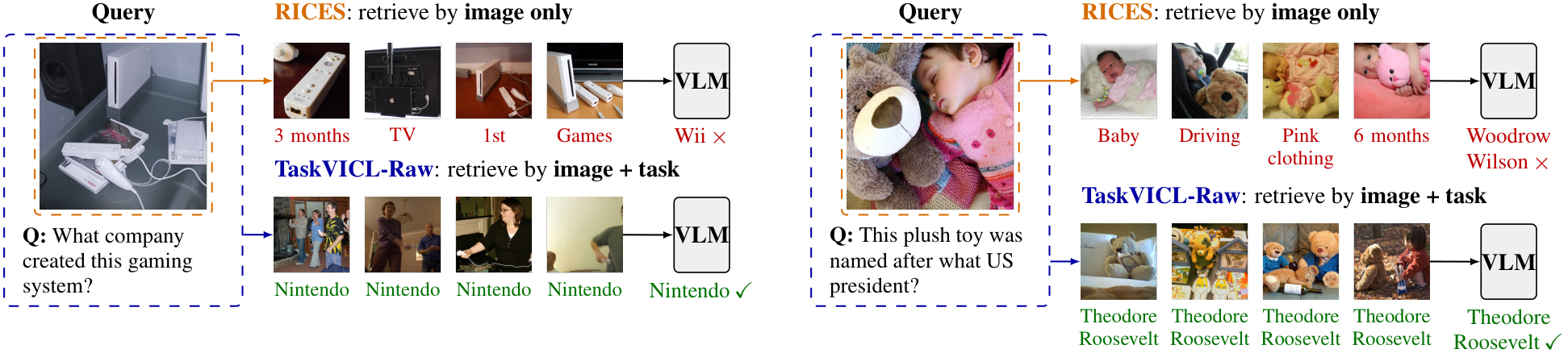}
\caption{OK-VQA examples with InternVL. RICES retrieves visually similar images whose questions differ from the query, while TaskVICL-Raw retrieves demonstrations asking the same question.}
\label{fig:vqa_examples}

\end{figure*}

We cross the extraction source (Task or Generic) with the answers shown for the retrieved examples, either reference answers or the VLM's own zero-shot predictions (Model), while the retrieved examples stay fixed. On OK-VQA (Table~\ref{tab:answers}), reference answers add 6.14 points under Task retrieval with interval $[5.16,7.08]$ and only 0.80 under Generic retrieval with interval $[0.19,1.41]$. Figure~\ref{fig:vqa_examples} illustrates this on two OK-VQA queries. Image-only retrieval returns visually similar images whose answers address other questions, whereas Raw returns examples that ask the same question, so their reference answers apply directly.

\begin{table}[t]
\centering\small
\caption{Demonstration answers with fixed retrieved examples on Qwen. Model shows the VLM's own zero-shot predictions as demonstration answers, and Reference shows bank reference answers. OK-VQA uses Raw with Task or Generic extraction text. On CUB, Initially wrong covers the 3640 of 5794 queries Qwen misclassifies without demonstrations, and RICES uses CLIP ViT-B/32.}
\label{tab:answers}\label{tab:corrective}
\setlength{\tabcolsep}{4pt}
\renewcommand{\arraystretch}{0.85}
\setlength{\aboverulesep}{0.2ex}
\setlength{\belowrulesep}{0.2ex}
\begin{tabular*}{\linewidth}{@{\extracolsep{\fill}}llrr@{}}
\toprule
Setting & Retrieval & Model & Reference\\
\midrule
\multirow{2}{*}{OK-VQA} & Raw, Task & 54.61 & \textbf{60.75}\\
                        & Raw, Generic & 53.27 & 54.07\\
\midrule
\multirow{2}{*}{CUB, all} & RICES & 36.18 & 71.21\\
                          & Raw & 37.73 & \textbf{81.50}\\
\midrule
\multirow{2}{*}{CUB, initially wrong} & RICES & 7.88 & 63.30\\
                                      & Raw & 4.62 & \textbf{74.92}\\
\bottomrule
\end{tabular*}

\end{table}

We then ask whether reference answers help recover queries the VLM gets wrong on its own. On CUB, Qwen first predicts every bank and query label without demonstrations. Model shows these predictions as demonstration answers, and Reference replaces the wrong ones with reference labels. Reference answers add 43.77 points for Raw and 35.04 for RICES (Table~\ref{tab:corrective}). On the 3640 initially wrong queries, Raw rises from 4.62 to 74.92 and RICES from 7.88 to 63.30. When the two retrievers are matched on the number of wrong demonstration answers, Raw still gains 10.23 points more than RICES, with interval $[8.00,12.52]$ (Appendix~\ref{sec:supp_source}).

Raw retrieves well even where the model itself answers poorly. Qwen labels only 37.18\% of CUB queries correctly without demonstrations, yet 72.20\% of Raw's retrieved examples share the query class, against 46.62\% for RICES (Appendix~\ref{sec:supp_additional}). Images of one class thus receive similar states even when the model misnames them, and the reference answers supply the correct name.

\begin{figure*}[t]
\centering
\includegraphics[width=\textwidth]{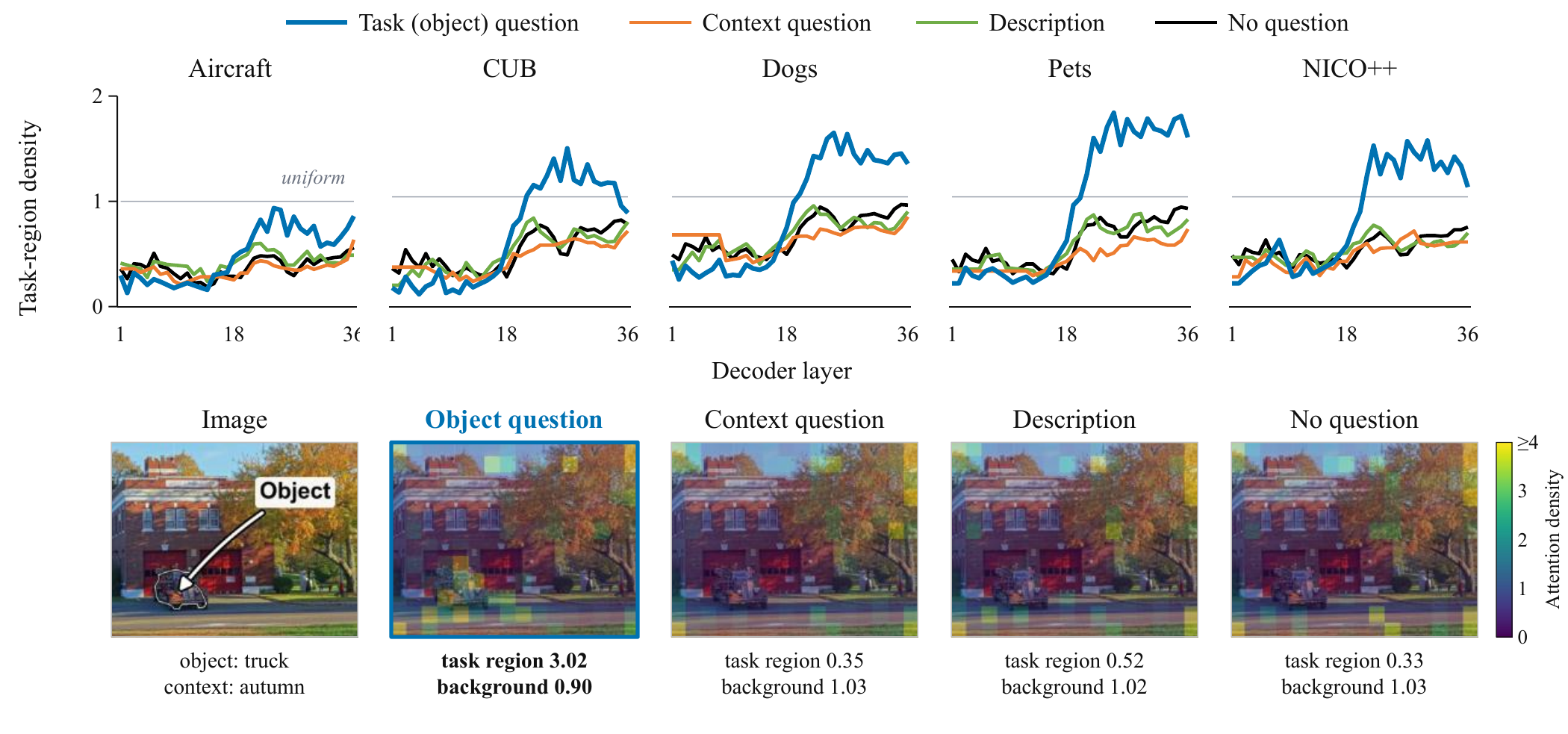}
\caption{Attention at the final input position. Top: mean task-region density per layer; Task is the object question, and grey lines mark uniform density. Bottom: a NICO++ example averaged over 36 layers, selected after inspecting maps.}
\label{fig:reading}

\end{figure*}

\section{Task-Conditioned Visual Attention}
\label{sec:explanation}
To see how the task text changes what the extraction position reads from the image, we examine attention at the final input position $T$ used for retrieval. We sample 300 images from each of Aircraft, CUB, Dogs, Pets and NICO++ before extracting attention. Each image receives four prompts. Task uses the original instruction (the object question on NICO++), Context asks about the image context, Description requests a generic description and No question omits the question.

We average attention from $T$ to the $V$ image tokens over heads and layers and normalize it to $p_v$ over image tokens. The task region covers the classified object, that is the aircraft, bird, dog, pet or NICO++ object, and stays fixed across prompts in original image coordinates. For a region with token coverage $r_v\in[0,1]$, the density $D_R=\sum_v r_vp_v/(\sum_v r_v/V)$ divides the region's share of image attention by its share of area, so uniform attention gives one (Appendix~\ref{sec:supp_attention}).

\begin{table}[t]
\centering\small
\caption{Mean task-region density with attention averaged over 36 layers. $\Delta$ is Task $-$ Context with paired image-bootstrap 95\% intervals.}
\label{tab:reading_density}
\setlength{\tabcolsep}{4pt}
\renewcommand{\arraystretch}{0.85}
\setlength{\aboverulesep}{0.2ex}\setlength{\belowrulesep}{0.2ex}
\begin{tabular*}{\linewidth}{@{\extracolsep{\fill}}lrrrrl@{}}
\toprule
Dataset & Task & Context & Descr. & No quest. & \multicolumn{1}{c}{$\Delta$}\\
\midrule
Aircraft & \textbf{0.585} & 0.349 & 0.429 & 0.375 & $+0.24$ {\scriptsize$[0.22, 0.25]$}\\
CUB      & \textbf{0.844} & 0.432 & 0.529 & 0.492 & $+0.41$ {\scriptsize$[0.39, 0.44]$}\\
Dogs     & \textbf{1.076} & 0.569 & 0.666 & 0.643 & $+0.51$ {\scriptsize$[0.45, 0.57]$}\\
Pets     & \textbf{1.056} & 0.451 & 0.576 & 0.549 & $+0.61$ {\scriptsize$[0.56, 0.66]$}\\
NICO++   & \textbf{0.970} & 0.501 & 0.539 & 0.523 & $+0.47$ {\scriptsize$[0.41, 0.54]$}\\
\bottomrule
\end{tabular*}

\end{table}

Task gives the highest task-region density on all five datasets (Table~\ref{tab:reading_density}). Relative to the context question, it raises density by 0.24 to 0.61 with every interval excluding zero, and it does so on at least 95\% of the images in each dataset. Description and No question also stay well below Task, so the shift requires the task text. Figure~\ref{fig:reading} shows that the shift emerges after about layer 18, where the Task curve rises above uniform density on CUB, Dogs, Pets and NICO++. We use these measurements to describe internal processing and measure demonstration utility through predictions \cite{jainwallace,adebayo}.

\section{Conclusion}
We presented TaskVICL, which selects visual in-context demonstrations with task-conditioned decoder states of the frozen target VLM. Its training-free Raw variant compares all decoder layers before generation. Across two VLMs and eight datasets, Raw has the highest average among training-free retrievers, and on NICO++ it adapts its demonstrations to the question asked. Ablations show that task text at extraction underlies most gains and that bank reference answers correct many initial errors; task text also draws more attention to the classified object. A projection learned from bank labels adds classification gains at a tenth of the storage, and unsupervised PCA keeps question answering accuracy at the same cost.

\paragraph{Limitations.}
TaskVICL reads hidden states and therefore applies only to open-weight VLMs. Raw stores all layer states (360 KiB per Qwen bank entry), and a new target model or label set requires re-extracting the bank. Retrieval quality also depends on the target model: on InternVL Aircraft and CUB, where zero-shot accuracy is below 15\%, Raw retrieves about as many same-class examples as RICES. Our largest evaluated model is Gemma-3-12B-IT, and our intervals cover query sampling but not variation from bank splits, prompt templates or demonstration order.

\paragraph{Acknowledgments.}
This research/project is supported by the National Research Foundation, Singapore under its National Large Language Models Funding Initiative (AISG Award No: AISG-NMLP-2024-004). Any opinions, findings and conclusions or recommendations expressed in this material are those of the author(s) and do not reflect the views of National Research Foundation, Singapore.

\clearpage
\section*{LLM Usage Statement}
ChatGPT assisted with editing the manuscript, revising its language, checking related-work descriptions and editing the LaTeX source. Claude (Anthropic) assisted with revising and condensing the text, editing the LaTeX source and TikZ figures, checking numerical consistency across tables, and simulating reviewer feedback on drafts. ChatGPT also helped organize the experimental discussion and summarize the paired corrective-effect analysis. Reported quantitative results come from the authors' experimental records. Codex assisted with approximate image-only subject outlines for the NICO++ attention study before attention maps were extracted. Qwen3-VL-4B and InternVL3.5-4B are the target models evaluated in the experiments.

{
\small
\bibliographystyle{ieeenat_fullname}
\bibliography{references}
}
\clearpage
\appendix
\section*{Supplementary Material}
\pdfbookmark[0]{Supplementary Material}{supplementary}
\setcounter{table}{0}
\setcounter{figure}{0}
\setcounter{equation}{0}
\renewcommand{\thetable}{S\arabic{table}}
\renewcommand{\thefigure}{S\arabic{figure}}
\renewcommand{\theequation}{S\arabic{equation}}
\renewcommand{\theHtable}{supp.\arabic{table}}
\renewcommand{\theHfigure}{supp.\arabic{figure}}
\renewcommand{\theHequation}{supp.\arabic{equation}}
\section{Implementation Details}
\label{sec:supp_implementation}

\paragraph{State extraction.}
TaskVICL extracts the state at the final input position of each decoder layer, including the assistant prefix, before any answer is generated. Images go through the target VLM's own processor. Qwen uses processor pixel bounds of 65,536 and 16,777,216, and its classification messages carry $224\times224$ image-size hints. InternVL uses native dynamic tiling with up to 12 tiles plus a thumbnail, each $448\times448$ with 256 visual tokens. Both models run in BF16.

\paragraph{Prediction.}
Classification, including zero-shot, uses legal-label decoding. Question answering uses greedy decoding with at most 32 new tokens. The four demonstrations appear in ascending similarity, with the nearest last. Without demonstrations, InternVL often appends the option text to the ScienceQA option letter, so 31.9\% of its zero-shot ScienceQA outputs are invalid, while all four-shot ScienceQA runs have at least 99.7\% valid outputs. The equal InternVL OK-VQA scores of RICES and Random-4 (47.62) come from two separate runs with different demonstrations.

\paragraph{Uncertainty.}
Seed 73 fixes the sampling of Random-4 and the training of Proj and DeTriever, and the other retrievers are deterministic. Paired intervals use 2000 bootstrap resamples of the same query units with seed 73. The CUB answer intervention and its matched analyses use 20000 resamples. NICO++ resamples the 1080 images with both questions kept together. These intervals describe query uncertainty for fixed runs, and Sec.~\ref{sec:supp_additional} reports variation across seeds.

\section{Baseline Details}
\label{sec:supp_baselines}

All baselines receive the image and the original task text, search the same complete bank, return four demonstrations in ascending similarity and use the common final prompt with reference answers. Query reference answers never enter representation extraction or retriever training.

\paragraph{RICES and image encoders.}
RICES ranks bank images by cosine similarity of L2-normalized FP32 image features. RICES without a qualifier denotes the CLIP ViT-B/32 variant throughout. CLIP ViT-B/32 and ViT-L/14 use the OpenAI checkpoints at $224\times224$. DINOv3 uses the frozen ViT-B/16 model at its native 256-pixel resolution with mean patch features. The vision-tower control mean-pools the output tokens of Qwen3-VL-4B's own vision encoder under its native processor. On NICO++, the four demonstrations have distinct images.

\paragraph{VLM2Vec.}
We use \nolinkurl{TIGER-Lab/VLM2Vec-Full} with frozen weights, its Phi vision processor and 16 crops. The input combines the image with a fixed representation instruction and the complete task text, including the label list for classification and the question and options for question answering. The normalized 3072-dimensional final state at the last input position ranks the bank by cosine similarity.

\paragraph{Caption selectors.}
Qwen3-VL-4B generates one question-independent description per image with at most 64 new tokens, computed offline. Each description is joined with the original task text. BGE-M3 embeds this text into a normalized 1024-dimensional CLS feature with maximum length 8192, and BM25 ranks it with $k_1=1.2$ and $b=0.75$. The banks contain 3760 DTD, 5994 CUB and 9009 OK-VQA entries.

\paragraph{GPT-MM.}
GPT-MM combines task-aware retrieval with label-induced reasoning \cite{gptmm}. We evaluate its retrieval component with the target VLM as the representation model and keep the shared final prompt without generated reasoning. The frozen VLM first answers every bank example and query greedily without demonstrations, with at most 12 new tokens for classification and 32 for question answering. Let $m$ be the number of generated tokens before the first end-of-sequence token and $T$ the number of available hidden-state steps. We take the final-layer state at the last position of step $\min(\max(1,m),T-1)$, or the last step if generation reaches the cap. These 2560-dimensional vectors are normalized and ranked by cosine similarity without fitted parameters.

\paragraph{DeTriever.}
DeTriever learns a weighted combination of decoder-layer transformations for NL2SQL retrieval \cite{detriever}. We apply it to the frozen final-input states of decoder layers $\mathcal I=\{1,6,11,16,21,26,31,36\}$. Each layer has its own MLP $f_\ell$ with widths $2560\to1024\to1024\to512$ and ReLU after the first two maps, and learned logits $a_\ell$ give the layer weights,
\begin{equation}
\begin{aligned}
e(x,t) &= \nu\Bigl(\sum_{\ell\in\mathcal I}\alpha_\ell f_\ell(h_\ell(x,t))\Bigr),\\
\alpha_\ell &= \frac{\exp(a_\ell)}{\sum_{j\in\mathcal I}\exp(a_j)}.
\end{aligned}
\end{equation}
For the training target, we append each bank example's reference answer as the assistant response and take the teacher-forced final-layer state at the last chat position. Dot products between these vectors rank the other bank entries for each anchor, and the 40 most and 100 least similar entries form its positives and negatives. A supervised contrastive loss trains the head for 10000 steps with batch size 64, temperature 0.07, AdamW with learning rate $10^{-4}$, weight decay 0.01 and coefficients $(0.9,0.98)$, refreshing a cache of candidate embeddings every 100 steps. A separate head is trained per model and dataset on the bank alone. Queries are encoded from the image and task text before generation.

\paragraph{CIRCLES.}
CIRCLES generates attribute and counterfactual descriptions with the target VLM, using at most 512 new tokens per auxiliary generation, and encodes images and text with CLIP ViT-g/14, truncating text to 77 tokens \cite{circles}. It retains its two image-based and two counterfactual-based retrieval arms. Images are distinct within each arm, and an image retrieved by both arms appears twice.

\paragraph{MMICES.}
We follow Algorithm~1 of MMICES \cite{mmices}. It preselects the $K=32$ bank entries with the highest visual similarity to the query and selects $N=4$ of them by text similarity, placing the most similar last. Visual features are the visual tokens that the target VLM's own vision encoder passes to the language model, mean-pooled and L2-normalized. Qwen uses its native resolution, and InternVL uses a single $448\times448$ view without tiling. The original method runs on OpenFlamingo, whose vision encoder is CLIP, so this choice carries its design over to the evaluated VLMs. Text similarity is the cosine similarity of CLIP ViT-B/32 text features with 77-token truncation. The text contains only the question and never an answer. For classification it is the task instruction, which is identical for all entries. On NICO++ it is the object or context question, and the two questions differ from the second token. For ScienceQA and OK-VQA it is the question sentence without instruction, context or options, and every such text fits within 77 tokens. Ties in text similarity keep the visual order. On the five single-task datasets the reranking is therefore constant, and the selected examples equal the four visual nearest neighbors for every query. Qwen DTD accuracy is accordingly 82.61, against 82.66 for the vision-tower RICES control. On NICO++ every selected example shares the query's question type. Prediction uses the four-shot prompt and decoding of the main results. Table~\ref{tab:supp_mmices} gives paired intervals for Raw against MMICES.

\begin{table}[H]\centering\small
\caption{Raw minus MMICES, in points with paired 95\% intervals. NICO++ resamples images.}
\label{tab:supp_mmices}
\setlength{\tabcolsep}{2pt}
\begin{tabular*}{\linewidth}{@{\extracolsep{\fill}}lcc@{}}\toprule
Dataset & Qwen3-VL-4B & InternVL3.5-4B\\\midrule
NICO++ object & $+0.46$ $[-0.56,1.48]$ & $+1.57$ $[0.46,2.78]$\\
NICO++ context & $+2.13$ $[0.09,4.17]$ & $+2.69$ $[1.37,4.01]$\\
DTD & $+1.44$ $[0.16,2.71]$ & $0.00$ $[-1.70,1.70]$\\
Aircraft & $+8.49$ $[6.84,10.17]$ & $+8.88$ $[7.11,10.68]$\\
CUB & $+3.16$ $[2.11,4.21]$ & $+25.54$ $[24.02,27.01]$\\
Dogs & $+5.82$ $[4.95,6.70]$ & $+22.39$ $[21.18,23.59]$\\
Pets & $+3.33$ $[2.51,4.14]$ & $+23.85$ $[22.21,25.57]$\\
ScienceQA & $+3.92$ $[2.68,5.11]$ & $+0.59$ $[-0.45,1.54]$\\
OK-VQA & $+3.78$ $[2.68,4.91]$ & $+2.85$ $[1.88,3.80]$\\
\bottomrule\end{tabular*}\end{table}

\section{Projection Variants}
\label{sec:supp_projection}

\paragraph{Supervised projection.}
Proj applies one bias-free linear map to every normalized layer state, with $2560\times256=655360$ parameters for Qwen. The contrastive objective sums the exponentiated similarities to all same-group examples inside the logarithm, and the denominator contains every other batch example. Group slots may repeat when fewer than 16 eligible groups exist. Classification holds out ten bank examples per class for checkpoint selection. A supervised variant for question answering groups bank questions by normalized training answer and keeps groups with at least 14 questions, ten of which form the development split. OK-VQA then yields only five groups (kite, travel, ski, mountain and eat), with 83 questions in total, 33 for training and 50 for development. This weakly supervised Proj gives the OK-VQA Proj entries in Secs.~F and H, and Table~\ref{tab:main} uses PCA for question answering.

\paragraph{PCA for question answering.}
PCA fits a 256-dimensional basis to bank states without answers, and projected layer states are normalized and averaged uniformly as in Proj. PCA minus Raw is $+0.15$ $[-0.49,0.79]$ on Qwen ScienceQA, $+0.04$ $[-0.63,0.71]$ on Qwen OK-VQA, $-0.15$ $[-0.79,0.40]$ on InternVL ScienceQA and $+0.30$ $[-0.31,0.92]$ on InternVL OK-VQA. PCA therefore keeps Raw's accuracy at a tenth of the storage.

\paragraph{Learned layer weights.}
Learning layer weights on top of Proj changes Qwen accuracy by at most 0.59 points across the five classification datasets and the two question answering datasets. On InternVL it adds 0.27 to 1.78 points on Aircraft, CUB, Dogs and Pets. We keep uniform weights for simplicity.

\paragraph{Projection forms.}
We compare projection forms on Qwen DTD and CUB with the same cached states, development split (3290 and 470 DTD images, 3994 and 2000 CUB images), uniform averaging and final prompts. Gaussian entries have variance $1/p$, PCA pools normalized training-bank states across layers without whitening, and the bias-free MLP has widths $2560\to1024\to256$. Learned forms share the objective and 1500 AdamW steps with learning rate 0.0003 and weight decay 0.01, refitted separately from the main-table checkpoints. The paired 95\% intervals for Linear minus X are $[1.12,3.46]$, $[0.74,3.24]$ and $[-0.48,2.02]$ on DTD and $[1.19,2.83]$, $[2.05,3.73]$ and $[0.02,1.43]$ on CUB, for Gaussian, PCA and MLP respectively. Table~\ref{tab:supp_projection_forms} reports development label precision at four. The linear map exceeds the Gaussian and PCA projections by 1.97 to 2.89 points, so the gain of Proj stems from bank supervision and dimensionality reduction alone does not explain it. The MLP adds 2.2M parameters without improving over the linear map.

\begin{table}[H]
\centering\small
\caption{Projection forms at $p=256$ on Qwen. Dev.\ is development label precision at four on held-out bank examples. Linear $-$ X is the accuracy difference in points.}
\label{tab:supp_projection_forms}
\setlength{\tabcolsep}{3pt}
\begin{tabular*}{\linewidth}{@{\extracolsep{\fill}}lrrrr@{}}
\toprule
 & \multicolumn{2}{c}{DTD} & \multicolumn{2}{c}{CUB}\\
\cmidrule(lr){2-3}\cmidrule(l){4-5}
Projection & Dev. & Lin.\ $-$ X & Dev. & Lin.\ $-$ X\\
\midrule
Gaussian & 78.72 & $+2.29$ & 68.44 & $+2.03$\\
PCA & 78.51 & $+1.97$ & 67.11 & $+2.89$\\
Shared MLP & \textbf{84.89} & $+0.80$ & 72.41 & $+0.75$\\
Shared linear & 83.51 & -- & \textbf{74.12} & --\\
\bottomrule
\end{tabular*}
\end{table}

\paragraph{Width.}
Table~\ref{tab:supp_projection_widths} varies the linear output width from 32 to 512 with everything else fixed. Accuracy changes by less than one point across widths, so even $p=32$ (1.25\% of the Raw cache) retains most of the gain. The main paper keeps $p=256$, fixed before this study.

\begin{table*}[!tbp]
\centering\small
\caption{Shared linear width on Qwen. Acc.\ and Dev.\ (development precision at four) are percentages. $\Delta$ is accuracy minus the $p=256$ run with paired 95\% intervals. Cache is the FP32 feature size relative to Raw.}
\label{tab:supp_projection_widths}
\setlength{\tabcolsep}{5pt}
\begin{tabular}{rrrcrrcr}
\toprule
 & \multicolumn{3}{c}{DTD} & \multicolumn{3}{c}{CUB} & \\
\cmidrule(lr){2-4}\cmidrule(lr){5-7}
$p$ & Acc. & Dev. & $\Delta$ & Acc. & Dev. & $\Delta$ & Cache (\%)\\
\midrule
32 & 85.48 & 83.19 & $-0.16$ $[-0.96,0.64]$ & 82.45 & 73.38 & $-0.71$ $[-1.33,-0.05]$ & 1.25\\
64 & 85.43 & \textbf{83.56} & $-0.21$ $[-0.96,0.53]$ & \textbf{83.26} & 74.30 & $+0.10$ $[-0.47,0.67]$ & 2.50\\
128 & 85.21 & 83.46 & $-0.43$ $[-1.12,0.32]$ & 82.90 & \textbf{74.45} & $-0.26$ $[-0.79,0.31]$ & 5.00\\
256 & 85.64 & 83.51 & -- & 83.15 & 74.12 & -- & 10.00\\
512 & \textbf{85.85} & 83.46 & $+0.21$ $[-0.43,0.80]$ & 82.88 & 74.16 & $-0.28$ $[-0.76,0.21]$ & 20.00\\
\bottomrule
\end{tabular}
\end{table*}

\section{Additional Results}
\label{sec:supp_additional}

\paragraph{Repeated runs.}
Table~\ref{tab:supp_three_seeds} repeats the Qwen classification evaluation with seeds 73, 74 and 75. RICES, GPT-MM, CIRCLES and Raw retrieve deterministically, so their spread reflects runtime differences between resumed runs. For DeTriever and Proj it also includes the training seed. Standard deviations stay below 0.5 points for Raw and Proj, and their ranking against the baselines is unchanged on every dataset.

\begin{table*}[!tbp]
\centering\small
\caption{Qwen classification accuracy over seeds 73, 74 and 75, as mean $\pm$ sample standard deviation.}
\label{tab:supp_three_seeds}
\setlength{\tabcolsep}{5pt}
\begin{tabular}{lccccc}
\toprule
Method & DTD & Aircraft & CUB & Dogs & Pets\\
\midrule
RICES & $78.71 \pm 0.08$ & $59.71 \pm 0.05$ & $71.23 \pm 0.03$ & $74.25 \pm 0.07$ & $91.59 \pm 0.11$\\
GPT-MM & $81.56 \pm 0.06$ & $68.85 \pm 0.49$ & $72.78 \pm 0.24$ & $78.67 \pm 0.02$ & $93.05 \pm 0.33$\\
DeTriever & $78.09 \pm 0.68$ & $68.29 \pm 0.47$ & $70.69 \pm 1.76$ & $77.69 \pm 0.89$ & $92.88 \pm 0.43$\\
CIRCLES & $80.83 \pm 0.13$ & $67.53 \pm 0.03$ & $81.94 \pm 0.04$ & $79.33 \pm 0.03$ & $93.29 \pm 0.03$\\
TaskVICL-Raw & $84.01 \pm 0.11$ & $73.95 \pm 0.10$ & $81.45 \pm 0.18$ & $83.05 \pm 0.10$ & $94.90 \pm 0.12$\\
TaskVICL-Proj & $85.43 \pm 0.44$ & $78.62 \pm 0.14$ & $83.42 \pm 0.18$ & $85.05 \pm 0.13$ & $95.38 \pm 0.21$\\
\bottomrule
\end{tabular}
\end{table*}

\paragraph{Number of demonstrations.}
Table~\ref{tab:supp_kshot} varies the number of demonstrations on Qwen DTD, CUB and OK-VQA with $k\in\{1,4,8\}$. The $k$ nearest bank examples enter the prompt in ascending similarity order, with the nearest last. RICES uses CLIP ViT-B/32. The DTD and CUB RICES results use the main-table rankings. On OK-VQA, the reported four-shot scores are 60.75 for Raw and 57.52 for RICES, compared with 60.75 and 57.63 in the main table. Within each dataset and method, the one-, four- and eight-shot comparisons shown here use the corresponding ranking. OK-VQA uses the original prompt with only the number of answered examples changed, and the official ten-annotator consensus score.

\begin{table}[H]\centering\small
\caption{Qwen results with $k$ demonstrations on DTD (1{,}880 queries), CUB (5{,}794) and OK-VQA (5{,}046 queries, consensus score). Raw $-$ RICES uses paired 95\% query-bootstrap intervals.}
\label{tab:supp_kshot}
\setlength{\tabcolsep}{4pt}
\begin{tabular*}{\linewidth}{@{\extracolsep{\fill}}lrrl@{}}\toprule
$k$ & RICES & Raw & Raw $-$ RICES\\\midrule
\multicolumn{4}{@{}l}{\textit{DTD}}\\
1 & 73.83 & 82.71 & $+8.88$ $[7.18,10.80]$\\
4 & 78.72 & 84.04 & $+5.32$ $[3.77,6.97]$\\
8 & 80.37 & 84.15 & $+3.78$ $[2.29,5.27]$\\\midrule
\multicolumn{4}{@{}l}{\textit{CUB}}\\
1 & 58.66 & 75.61 & $+16.95$ $[15.57,18.29]$\\
4 & 71.21 & 81.50 & $+10.29$ $[9.11,11.41]$\\
8 & 73.73 & 81.60 & $+7.87$ $[6.78,9.01]$\\\midrule
\multicolumn{4}{@{}l}{\textit{OK-VQA}}\\
1 & 56.50 & 58.99 & $+2.48$ $[1.34,3.63]$\\
4 & 57.52 & 60.75 & $+3.23$ $[2.12,4.35]$\\
8 & 58.26 & 61.22 & $+2.95$ $[1.76,4.02]$\\
\bottomrule\end{tabular*}\end{table}

Raw exceeds RICES at every evaluated $k$ on all three datasets, and all nine available paired intervals exclude zero. On classification, RICES benefits more from additional demonstrations. From one to eight shots, RICES gains 6.54 points on DTD and 15.07 on CUB, compared with 1.44 and 5.99 for Raw. A single Raw neighbor exceeds eight RICES neighbors by 2.34 points on DTD and 1.88 on CUB. The Raw advantage narrows with $k$ on classification but remains at 3.78 and 7.87 points with eight shots. On OK-VQA, both methods gain about two points from one to eight demonstrations, and the Raw advantage stays between 2.48 and 3.23 points.

\paragraph{Demonstration order.}
TaskVICL places the four demonstrations in ascending similarity order, with the nearest last. To test whether this convention matters, we keep the four Raw demonstrations of every query fixed and change only their order. Nearest first reverses the order. Random applies a uniformly sampled permutation per query with seed 73. We evaluate Qwen3-VL-4B-Instruct on all DTD and CUB queries with the main-table prediction settings. All three orders use the same re-extracted Raw rankings. Nearest last scores 83.67 on DTD and 81.20 on CUB, compared with 84.04 and 81.50 in the main table with the original states.

\begin{table}[H]
\centering\footnotesize
\caption{Qwen accuracy with the same four Raw demonstrations in different orders. Differences from nearest last are paired 95\% query-bootstrap intervals (2000 resamples, seed 73). DTD has 1880 queries and CUB 5794.}
\label{tab:supp_order}
\setlength{\tabcolsep}{2pt}
\begin{tabular*}{\linewidth}{@{\extracolsep{\fill}}lrr@{}}
\toprule
Order & DTD & CUB\\
\midrule
Nearest last (default) & 83.67 & 81.20\\
Nearest first & 84.36 & 81.39\\
Random & 83.78 & 80.95\\
\midrule
Nearest first $-$ last & $+0.69$ $[-0.16,1.54]$ & $+0.19$ $[-0.40,0.78]$\\
Random $-$ last & $+0.11$ $[-0.74,0.91]$ & $-0.26$ $[-0.88,0.38]$\\
\bottomrule
\end{tabular*}
\end{table}

All four order differences in Table~\ref{tab:supp_order} are within 0.69 points and their intervals include zero. Under the tested permutations, performance is not strongly dependent on nearest-last ordering.

\paragraph{Label precision of retrieved examples.}
For a classification query $q$ with label $y_q$ and demonstration labels $y_{q1},\ldots,y_{q4}$, label precision@4 is $100\sum_q c_q/(4Q)$ with $c_q=\sum_{j=1}^{4}\mathbf{1}[y_{qj}=y_q]$. Query labels are used only for this diagnostic after retrieval, and repeated CIRCLES images count at every position. Table~\ref{tab:supp_label_precision} shows that Raw retrieves more same-class examples than RICES on all five Qwen datasets, and Proj raises precision further on every dataset for both models. On InternVL Aircraft and CUB, where zero-shot accuracy is only 14.04 and 14.93, Raw is close to RICES.

\begin{table*}[!tbp]
\centering\small
\caption{Label precision@4 (\%) with the mean number of same-class demonstrations in parentheses.}
\label{tab:supp_label_precision}
\setlength{\tabcolsep}{5pt}
\begin{tabular*}{\textwidth}{@{\extracolsep{\fill}}lrrrrr@{}}
\toprule
Method & DTD & Aircraft & CUB & Dogs & Pets\\
\midrule
\multicolumn{6}{l}{\textbf{Qwen3-VL-4B}}\\
\midrule
Random-4 & 2.05 (0.08) & 0.97 (0.04) & 0.50 (0.02) & 0.73 (0.03) & 2.87 (0.11)\\
RICES & 60.56 (2.42) & 31.56 (1.26) & 46.62 (1.86) & 39.24 (1.57) & 65.71 (2.63)\\
GPT-MM & 76.61 (3.06) & 55.74 (2.23) & 58.13 (2.33) & 69.53 (2.78) & 88.54 (3.54)\\
DeTriever & 67.50 (2.70) & 57.21 (2.29) & 51.54 (2.06) & 66.22 (2.65) & 86.41 (3.46)\\
CIRCLES & 65.12 (2.60) & 33.15 (1.33) & 50.28 (2.01) & 45.72 (1.83) & 65.15 (2.61)\\
TaskVICL-Raw & 80.27 (3.21) & 64.48 (2.58) & 72.20 (2.89) & 76.81 (3.07) & 92.43 (3.70)\\
TaskVICL-Proj & \textbf{83.64} (3.35) & \textbf{71.95} (2.88) & \textbf{76.54} (3.06) & \textbf{80.09} (3.20) & \textbf{94.04} (3.76)\\
\midrule
\multicolumn{6}{l}{\textbf{InternVL3.5-4B}}\\
\midrule
Random-4 & 2.05 (0.08) & 0.97 (0.04) & 0.50 (0.02) & 0.73 (0.03) & 2.87 (0.11)\\
RICES & 60.56 (2.42) & 31.56 (1.26) & 46.62 (1.86) & 39.24 (1.57) & 65.71 (2.63)\\
GPT-MM & 62.38 (2.50) & 29.21 (1.17) & 34.18 (1.37) & 39.32 (1.57) & 67.97 (2.72)\\
DeTriever & 67.09 (2.68) & 34.19 (1.37) & 45.24 (1.81) & 55.18 (2.21) & 83.81 (3.35)\\
CIRCLES & 62.13 (2.48) & 31.51 (1.26) & 44.56 (1.77) & 40.32 (1.61) & 70.65 (2.83)\\
TaskVICL-Raw & 69.81 (2.79) & 32.85 (1.31) & 46.34 (1.85) & 53.47 (2.14) & 78.82 (3.15)\\
TaskVICL-Proj & \textbf{74.52} (2.98) & \textbf{39.69} (1.59) & \textbf{59.00} (2.36) & \textbf{61.99} (2.48) & \textbf{86.89} (3.48)\\
\bottomrule
\end{tabular*}
\end{table*}

\section{Label Voting over Retrieved Examples}
\label{sec:supp_voting}

This section is a supplementary diagnostic. TaskVICL is a demonstration retriever for VLM in-context learning, and the main paper evaluates it, like every baseline, through the predictions of the VLM that receives the demonstrations. Recent demonstration-selection methods for VLMs follow the same protocol and do not report a classifier that votes over the labels of the retrieved examples \cite{mmices,chen2025provoking,li2025taco,xu2025introspection,chen2025keco,codr,fu2026contextnav,gptmm,circles}. Baldassini et al.\ \cite{baldassini2024} compare RICES-based ICL with such voting and find that ICL often does not outperform it. We therefore report voting here to characterize the retrieved neighbors.

\paragraph{Protocol.}
Each query receives the majority reference label among a retriever's top four bank examples, without running the VLM. Ties go to the label of the nearest example. We evaluate the five Qwen single-task classification datasets with CLIP ViT-B/32, CLIP ViT-L/14 and TaskVICL-Raw neighbors.

\begin{table*}[!tbp]
\centering\small
\caption{Label voting over the top four retrieved examples (Vote) and four-shot ICL with the same neighbors (ICL) on Qwen. Values are accuracy percentages.}
\label{tab:supp_voting}
\setlength{\tabcolsep}{6pt}
\begin{tabular*}{\textwidth}{@{\extracolsep{\fill}}llrrrrrr@{}}
\toprule
Neighbors & Prediction & DTD & Aircraft & CUB & Dogs & Pets & Avg.\\
\midrule
RICES (CLIP ViT-B/32) & Vote & 69.04 & 39.99 & 58.61 & 51.48 & 74.71 & 58.77\\
RICES (CLIP ViT-L/14) & Vote & 75.37 & 55.75 & 77.13 & 71.32 & 88.09 & 73.53\\
TaskVICL-Raw & Vote & 82.13 & 70.33 & 78.41 & 81.75 & 93.10 & 81.14\\
\midrule
RICES (CLIP ViT-B/32) & ICL & 78.72 & 59.68 & 71.21 & 74.32 & 91.52 & 75.09\\
RICES (CLIP ViT-L/14) & ICL & 80.80 & 68.47 & 80.93 & 80.35 & 93.30 & 80.77\\
TaskVICL-Raw & ICL & \textbf{84.04} & \textbf{73.84} & \textbf{81.50} & \textbf{82.97} & \textbf{94.96} & \textbf{83.46}\\
\bottomrule
\end{tabular*}
\end{table*}

\paragraph{Raw states are a stronger retrieval representation.}
Table~\ref{tab:supp_voting} shows that voting over Raw neighbors exceeds voting over CLIP ViT-B/32 neighbors by 22.37 points on average and over CLIP ViT-L/14 neighbors by 7.61. Raw voting alone already exceeds ViT-B/32 RICES with four-shot ICL on all five datasets. The training-free states therefore place same-class examples much closer to the query than image encoders, which explains part of the ICL gains in Table~\ref{tab:main}.

\paragraph{ICL adds to the retrieved labels.}
With the same neighbors, four-shot ICL exceeds voting on every dataset and for every retriever, by 2.32 points on average for Raw. On NICO++, voting over Raw neighbors reaches 93.15 on object questions and 62.59 on context questions, against 94.54 and 63.70 with four-shot ICL. The VLM thus uses the demonstrations beyond their majority label, for example on queries whose neighbors disagree.

\paragraph{Question answering.}
Voting is undefined on ScienceQA and OK-VQA, whose options or open-ended answers differ across questions. There the comparison must go through the VLM, and Raw exceeds every baseline retriever on both datasets and both VLMs (Table~\ref{tab:main}). Relative to ViT-B/32 RICES, Raw gains 4.21 points on Qwen ScienceQA with interval $[3.02,5.45]$, 3.12 on Qwen OK-VQA with $[2.01,4.17]$ and 3.69 on InternVL OK-VQA with $[2.70,4.67]$. The InternVL ScienceQA difference of 0.79 has an interval including zero, $[-0.25,1.83]$.

\section{Layer and Representation Controls}
\label{sec:supp_layers}

\paragraph{Decoder layers.}
Early, Middle and Late use Qwen layers 1 to 12, 13 to 24 and 25 to 36, and Last uses layer 36. Table~\ref{tab:supp_layers} adds an untrained random projection and Proj at each depth. Later layers retrieve better than early ones, and Proj helps classification at every depth while a random projection of the same width changes little. All minus Last is $+2.32$ $[1.60,3.04]$ on DTD, $+1.57$ $[1.01,2.13]$ on CUB and $+1.70$ $[0.40,2.97]$ on OK-VQA. All minus Late is $+0.69$ $[0.02,1.30]$ on DTD, $+0.50$ $[0.08,1.09]$ on CUB and $-0.02$ $[-0.66,0.58]$ on OK-VQA.

\begin{table}[H]
\centering\small
\caption{Layer groups and projections on Qwen. Untrained is a fixed random projection with the width of Proj.}
\label{tab:supp_layers}
\setlength{\tabcolsep}{3pt}
\begin{tabular*}{\linewidth}{@{\extracolsep{\fill}}llrrr@{}}
\toprule
Layers & Metric & DTD & CUB & OK-VQA\\
\midrule
Last & Raw & 81.72 & 79.93 & 59.05\\
 & Untrained & 81.72 & 79.64 & 59.65\\
 & Proj & 85.05 & 83.76 & 60.42\\
\addlinespace[2pt]
Early & Raw & 71.01 & 32.05 & 58.07\\
 & Untrained & 71.54 & 32.12 & 57.97\\
 & Proj & 78.72 & 41.85 & 57.92\\
\addlinespace[2pt]
Middle & Raw & 83.24 & 78.05 & 59.22\\
 & Untrained & 82.98 & 78.03 & 59.26\\
 & Proj & 84.47 & 79.75 & 58.95\\
\addlinespace[2pt]
Late & Raw & 83.35 & 81.00 & 60.77\\
 & Untrained & 83.46 & 80.88 & 60.83\\
 & Proj & 85.48 & 84.17 & 60.59\\
\addlinespace[2pt]
All & Raw & 84.04 & 81.50 & 60.75\\
 & Untrained & 83.94 & 81.24 & 60.52\\
 & Proj & 85.90 & 83.36 & 60.62\\
\bottomrule
\end{tabular*}
\end{table}

\paragraph{Retrieval representations.}
Table~\ref{tab:supp_repr_pairs} gives paired differences between Raw and the image, embedding and caption retrievers on Qwen. Strong image encoders come close to Raw on one of DTD and CUB, yet every image encoder trails Raw on OK-VQA by more than three points. Over all eight datasets of Table~\ref{tab:main}, Raw exceeds CLIP ViT-L/14 RICES by 3.83 points on Qwen with interval $[3.38,4.27]$, and by 1.85 points on InternVL with interval $[1.41,2.26]$. Among the caption selectors, BGE-M3 exceeds BM25 on CUB by 10.37 points, and the two are on par on DTD and OK-VQA.

\begin{table*}[!tbp]\centering\small
\caption{Raw minus other retrieval representations on Qwen, in points with paired 95\% intervals.}
\label{tab:supp_repr_pairs}
\setlength{\tabcolsep}{8pt}
\begin{tabular}{lccc}\toprule
Representation & DTD & CUB & OK-VQA\\\midrule
RICES (CLIP ViT-B/32) & $+5.32$ $[3.77,6.97]$ & $+10.29$ $[9.11,11.41]$ & $+3.12$ $[2.01,4.17]$\\
RICES (CLIP ViT-L/14) & $+3.24$ $[1.76,4.73]$ & $+0.57$ $[-0.45,1.55]$ & $+3.58$ $[2.43,4.76]$\\
RICES (Qwen vision tower) & $+1.38$ $[0.05,2.71]$ & $+2.99$ $[1.90,4.06]$ & $+3.36$ $[2.26,4.50]$\\
VLM2Vec & $+2.39$ $[1.01,3.83]$ & $+11.24$ $[10.10,12.39]$ & $+0.06$ $[-0.83,0.97]$\\
Caption BGE-M3 & $+5.69$ $[4.15,7.23]$ & $+25.82$ $[24.49,27.15]$ & $+3.37$ $[2.27,4.45]$\\
Caption BM25 & $+5.16$ $[3.72,6.70]$ & $+36.19$ $[34.78,37.54]$ & $+3.71$ $[2.56,4.78]$\\
\bottomrule\end{tabular}\end{table*}

\section{NICO++ Paired Tasks}
\label{sec:supp_nico}

\paragraph{Construction.}
We sample from the official NICO++ public training set \cite{nico}, whose native labels cover 60 object classes and six contexts (autumn, dim, grass, outdoor, rock and water). We draw 84 images per object class and 840 per context, 5040 in total, after removing duplicate source paths. The bank holds 3960 images and the query set 1080, with three query images per object and context combination, split by image with seed 73. Images and labels are unchanged. We will release the bank and query lists, and the code upon acceptance of the paper. Each image supplies an object question with the 60-class list and a context question with the six-class list, giving 7920 bank entries and 2160 queries. All methods search the same mixed bank without filtering by question type, and labels are not edited.

\paragraph{Diagnostics.}
Plain RICES ranks images, and each retrieved image enters the prompt with its object question and the matching reference answer. Image label agreement at four is the share of retrieved images whose native object or context label matches the query image, inspected after retrieval. Neighbor overlap is the number of images shared by the object and context retrieval sets divided by four. The task-matched RICES control keeps the four RICES images and their order and supplies each with its question and reference answer for the query task, both of which are available for every bank image. It reuses the neighbors of each RICES encoder in the main results. Paired differences between Raw and task-matched RICES appear in Table~\ref{tab:supp_nico_accuracy}.

\begin{table}[H]
\centering\footnotesize
\caption{NICO++ Raw minus task-matched RICES, averaged over object and context questions, with paired image-bootstrap 95\% intervals. Full method accuracies are in the main paper.}
\label{tab:supp_nico_accuracy}
\setlength{\tabcolsep}{2pt}
\begin{tabular*}{\linewidth}{@{\extracolsep{\fill}}lcc@{}}
\toprule
Task-matched RICES & Qwen & InternVL\\
\midrule
CLIP ViT-B/32 & $+2.92$ $[1.62,4.21]$ & $+3.15$ $[1.90,4.44]$\\
CLIP ViT-L/14 & $+1.48$ $[0.19,2.73]$ & $+1.48$ $[0.19,2.78]$\\
DINOv3 & $+2.36$ $[1.11,3.66]$ & $+1.53$ $[0.23,2.82]$\\
\bottomrule
\end{tabular*}
\end{table}

\paragraph{Proj.}
NICO++ Proj uses the 66 groups formed by task and native label. Ten bank images per object class form the development split with both questions, the remaining 3360 images supply training entries, and final retrieval uses all 3960 bank images. 

\section{Extraction Text and Bank Answers}
\label{sec:supp_source}

\begin{figure}[t]
\centering\small
\includegraphics[width=\columnwidth]{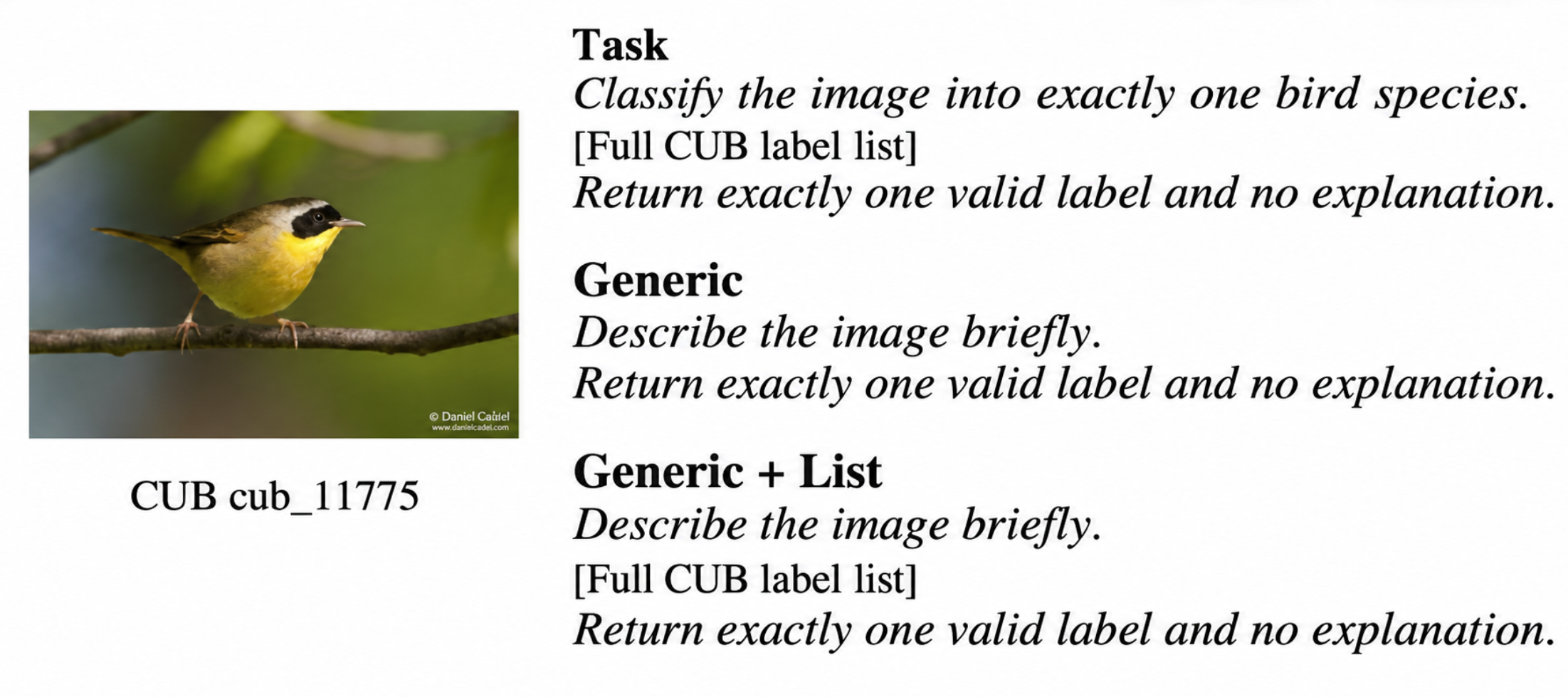}
\caption{Extraction inputs for one CUB image. The Task excerpt omits the full label list and formatting suffix for readability, and extraction uses the complete prompt. Both conditions lead to the original classification prompt at prediction.}
\label{fig:source_prompts}
\end{figure}

\paragraph{Extraction source on OK-VQA.}
Figure~\ref{fig:source_prompts} shows the Task and Generic extraction inputs. The Raw source-by-answer results are already reported in the main paper. Repeating the OK-VQA study with Proj, fitted separately for each source, gives 60.62 and 54.19 for Task retrieval with reference and model answers and 54.33 and 53.71 for Generic retrieval, closely following Raw.

\paragraph{CUB answer intervention.}
Qwen predicts every bank image and query of CUB (5994 and 5794) without demonstrations using legal-label decoding, which yields 2154 correct queries (37.18\%) and 3640 initially wrong ones. Raw and RICES keep their saved neighbors and order for every query. Model shows the independently predicted bank labels and Reference shows the reference labels, so the two conditions differ only where a bank prediction was wrong. The Reference predictions are those of Table~\ref{tab:main}, and their demonstration identities were checked query by query against the intervention records.

Table~\ref{tab:supp_corrective} shows that nearly all of the gain comes from initially wrong queries, where Raw gains 70.30 points and RICES 55.41. On initially correct queries, both changes have intervals including zero.

\begin{table*}[!tbp]
\centering\small
\caption{CUB answer intervention with 20000 paired query-bootstrap resamples. Rescued and Harmed count queries that switch from wrong to correct and from correct to wrong between Model and Reference.}
\label{tab:supp_corrective}
\setlength{\tabcolsep}{4pt}
\begin{tabular}{llrrrlrr}
\toprule
Queries & Retriever & Model & Reference & Gain & 95\% CI & Rescued & Harmed\\
\midrule
All (5794) & Raw & 37.73 & 81.50 & $+43.77$ & $[42.39,45.17]$ & 2672 & 136\\
All (5794) & RICES & 36.18 & 71.21 & $+35.04$ & $[33.62,36.45]$ & 2247 & 217\\
Initially wrong (3640) & Raw & 4.62 & 74.92 & $+70.30$ & $[68.82,71.81]$ & 2570 & 11\\
Initially wrong (3640) & RICES & 7.88 & 63.30 & $+55.41$ & $[53.74,57.12]$ & 2060 & 43\\
Initially correct (2154) & Raw & 93.69 & 92.62 & $-1.07$ & $[-2.46,0.32]$ & 102 & 125\\
Initially correct (2154) & RICES & 83.98 & 84.59 & $+0.60$ & $[-1.07,2.32]$ & 187 & 174\\
\bottomrule
\end{tabular}
\end{table*}

\paragraph{Matched subsets.}
Raw retrieves four independently mispredicted bank examples for 2620 queries, compared with 1908 for RICES, so its larger gain could reflect more corrected labels. We therefore restrict to the 1925 queries where both retrievers return the same nonzero number of mispredicted examples (116, 123, 239 and 1447 queries with one to four), 1714 of them initially wrong. Membership depends only on independent predictions and bank labels. Table~\ref{tab:supp_corrective_matched} shows that Raw still gains 10.23 points more than RICES on this subset, so the number of corrected labels alone does not explain the difference.

\begin{table*}[!tbp]
\centering\small
\caption{Matched CUB subsets. Gain diff.\ is the Reference minus Model gain of Raw minus that of RICES, with 20000 paired resamples.}
\label{tab:supp_corrective_matched}
\setlength{\tabcolsep}{4pt}
\begin{tabular}{lrrrrrrl}
\toprule
 & & \multicolumn{2}{c}{Raw} & \multicolumn{2}{c}{RICES} & &\\
Subset & $n$ & Model & Reference & Model & Reference & Gain diff. & 95\% CI\\
\midrule
Equal error count & 1925 & 11.12 & 79.06 & 10.39 & 68.10 & $+10.23$ & $[8.00,12.52]$\\
Initially wrong & 1714 & 2.86 & 78.65 & 3.33 & 67.33 & $+11.79$ & $[9.33,14.24]$\\
\bottomrule
\end{tabular}
\end{table*}

\begin{figure*}[t]
\centering
\includegraphics[width=\textwidth]{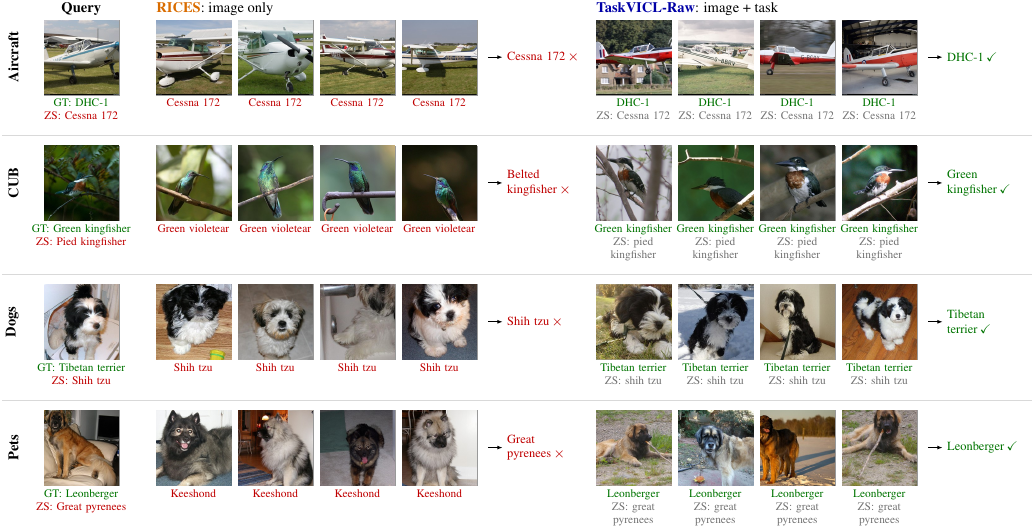}
\caption{RICES and TaskVICL-Raw on Qwen3-VL-4B with four demonstrations. GT is the query label and ZS the model's zero-shot prediction. Labels under each demonstration are its reference answer, green when it matches GT, and grey ZS gives the model's zero-shot prediction for that demonstration. Examples are selected among queries where zero-shot and RICES are wrong and Raw is correct.}
\label{fig:supp_cls_rices}
\end{figure*}

Figure~\ref{fig:supp_cls_rices} shows Qwen queries that the model misclassifies zero-shot, on which RICES stays wrong and Raw is correct. RICES returns visually similar images of other classes. Raw returns examples of the true class that the model itself labels as the same wrong class, and their reference answers correct the prediction.

\section{Attention Protocol}
\label{sec:supp_attention}

\paragraph{Images and prompts.}
We sample 300 images from each of Aircraft, CUB, Dogs, Pets and NICO++ before any attention is extracted. Each image receives the Task, Context, Description and No question prompts. The assistant prefix is present in all four, and the image tensors and visual token grids are identical across them.

\paragraph{Regions.}
The task region covers the classified object. Pets uses the released foreground trimap. Aircraft, CUB and Dogs use subject outlines drawn on the images, and NICO++ uses approximate subject outlines drawn from the images alone with assistance from Codex. All regions are fixed in original image coordinates before attention is extracted. It stays the same across the four prompts, and no image is removed based on attention. Each region maps to the token grid by fractional area coverage $r_v\in[0,1]$.

\paragraph{Attention and density.}
We reconstruct the attention row at the final input position $T$ from the saved normalized and position-encoded query and key tensors. The softmax covers all visible positions, including text. For selected layers $\mathcal S$, $H$ heads and $V$ image tokens,
\begin{equation}
a_v=\frac{1}{|\mathcal S|H}\sum_{\ell\in\mathcal S}\sum_{h=1}^{H}a_{\ell,h}(T,v),\qquad
p_v=\frac{a_v}{\sum_{u}a_u}.
\end{equation}
Raw weights are averaged across layers before normalization, which preserves differences in image attention mass across layers. Maps display $d_v=Vp_v$, and the regional density is $D_R=\sum_v r_vp_v/(\sum_v r_v/V)$, so uniform attention gives one. Early, Middle and Late average layers 1 to 12, 13 to 24 and 25 to 36. Intervals resample images with 2000 resamples and seed 73.

\begin{figure*}[t]
\centering
\includegraphics[width=\textwidth]{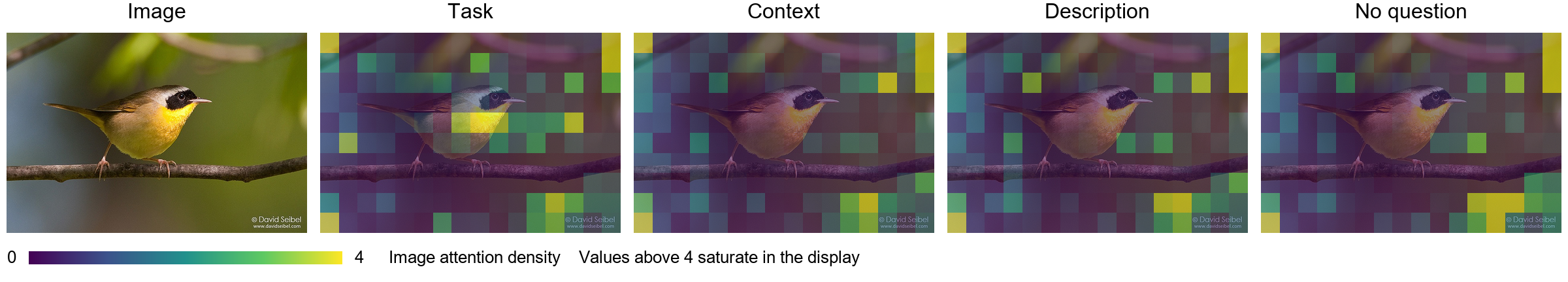}
\par\medskip
\includegraphics[width=\textwidth]{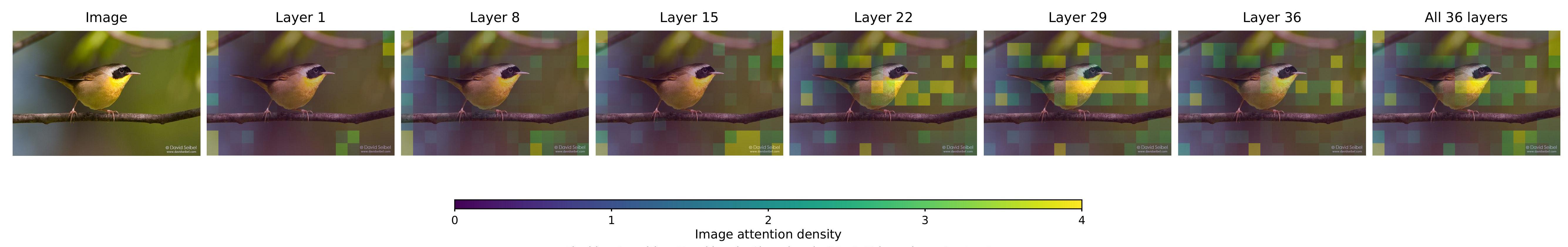}
\caption{Attention at the final input position for a CUB image. Top, the four prompts with attention averaged over 36 layers. Bottom, the Task prompt at six evenly spaced layers and the all-layer average. The Task prompt concentrates attention on the bird in the later layers. All panels share a density scale from zero to four.}
\label{fig:supp_cub_attention}
\end{figure*}

\paragraph{Display.}
Maps use nearest-neighbor cells on the native token grid overlaid on the image at opacity 0.55, with a shared viridis scale from zero to four. Larger values saturate in the display only, and all statistics use the unclipped values. The NICO++ truck example in the main paper and the CUB example in Figure~\ref{fig:supp_cub_attention} were chosen after inspecting maps for clear subjects. This choice does not affect the aggregate statistics.

\section{Computation and Storage}
\label{sec:supp_cost}

\paragraph{Cost model.}
Bank states are extracted once and cached. At query time, Raw needs one extra forward pass of the target VLM before generation and a bank comparison of cost $O(NLd)$ for $N$ entries, $L$ layers and width $d$. Proj applies the shared map and compares at $O(NLp)$. An FP32 Raw cache stores $4NLd$ bytes, which is 360 KiB per Qwen entry, and Proj at $p=256$ stores one tenth of that.

\paragraph{Measured latency.}
On one RTX 4090 with BF16, PyTorch 2.8.0 and CUDA 12.8, we time 100 CUB queries after 20 warm-up queries with cached bank features. Raw extraction takes 128.76 ms on average, search 2.41 ms and four-shot generation 461.05 ms, so retrieval adds about 28\% to generation time. Table~\ref{tab:supp_local_cost} compares methods in a second run on the same GPU over ten CUB queries after three warm-up queries, searching all 5994 bank entries. Retrieval time includes image loading, preprocessing, representation construction and search, and for CIRCLES also its attribute and counterfactual generation. This CIRCLES variant uses CLIP ViT-B/32 to match RICES. Model loading, bank construction and head training are excluded.

\begin{table}[H]
\centering\small
\caption{Pilot costs on ten CUB queries. Retrieval and generation are mean milliseconds. Peak is the maximum allocated GPU memory in GiB, including the VLM, any retrieval encoder and the bank cache. Bank is the FP32 feature size in MiB.}
\label{tab:supp_local_cost}
\setlength{\tabcolsep}{2.5pt}
\begin{tabular*}{\linewidth}{@{\extracolsep{\fill}}lrrrr@{}}
\toprule
Method & Retrieval & Generation & Peak & Bank\\
\midrule
RICES & 7.68 & 671.61 & 9.44 & 11.71\\
Raw & 267.35 & 674.92 & 10.91 & 2107.27\\
Proj ($p=256$) & 255.10 & 660.46 & 9.05 & 210.73\\
Proj ($p=32$) & 255.45 & 662.85 & 8.86 & 26.34\\
CIRCLES (ViT-B/32) & 10682.75 & 693.03 & 9.44 & 23.41\\
\bottomrule
\end{tabular*}
\end{table}

\section{Qwen3-VL-8B-Instruct}
\label{sec:supp_qwen8b}
To test whether gains persist at a larger scale within the Qwen family, we evaluate Qwen3-VL-8B-Instruct on NICO++, DTD and OK-VQA. To fit a single 16\,GB GPU, the language decoder is quantized to 4-bit NF4 with BF16 computation, while the vision encoder and output head remain in BF16. As in the Qwen3-VL-4B experiments, we use seed 73, complete banks and test sets, four demonstrations, $224\times224$ classification images and native-resolution NICO++ and OK-VQA images. NICO++ and DTD use legal-label decoding for all methods, including zero-shot. OK-VQA uses greedy decoding and ten-answer consensus scoring. Raw uses the final-input-position states of all 36 decoder layers ($36\times4096$), and RICES uses CLIP ViT-B/32. Paired intervals follow Sec.~\ref{sec:supp_implementation}, with image clusters for NICO++.

\begin{table*}[!t]
\centering\small
\caption{Qwen3-VL-8B-Instruct (4-bit NF4 decoder) with zero-shot and four-shot results. NICO++ reports object, context and mean accuracy. Bold marks the best score per column.}
\label{tab:supp_qwen8b}
\setlength{\tabcolsep}{8pt}
\begin{tabular}{lccccc}
\toprule
& \multicolumn{3}{c}{NICO++} & & \\
\cmidrule(lr){2-4}
Method & Obj. & Ctx. & Avg. & DTD & OK-VQA\\
\midrule
Zero-shot & 92.22 & 41.48 & 66.85 & 67.55 & 60.51\\
RICES & 93.06 & 50.37 & 71.71 & 80.32 & 57.34\\
Raw & \textbf{93.89} & \textbf{64.63} & \textbf{79.26} & \textbf{83.99} & \textbf{61.15}\\
\midrule
Raw $-$ RICES & $+0.83$ & $+14.26$ & $+7.55$ & $+3.67$ & $+3.80$\\
95\% CI & $[-0.19,1.85]$ & $[11.67,16.85]$ & $[6.11,8.94]$ & $[2.23,5.16]$ & $[2.67,4.95]$\\
\bottomrule
\end{tabular}
\end{table*}

Table~\ref{tab:supp_qwen8b} shows that Raw exceeds RICES in every column. All paired intervals exclude zero except NICO++ object, where zero-shot already reaches 92.22. The NICO++ mean gain of 7.55 points comes mainly from context questions ($+14.26$). On OK-VQA, zero-shot scores 60.51. RICES lowers the score by 3.17 points, while Raw scores 0.64 points above zero-shot. Relative to zero-shot, Raw gains 12.41 points on NICO++ and 16.44 on DTD. The quantized 8B experiment tests robustness across model sizes. The precision settings differ from those of the main experiments.

\section{Zero-shot Belief Retrieval}
\label{sec:supp_belief}

\paragraph{Representation.}
Every bank entry and query uses exactly the input of Raw, the same image and task prompt with the complete label list. For each legal label we compute the zero-shot log-likelihood of the label tokens followed by the end token. Labels are scored with and without a leading space, and the two are merged by log-sum-exp, matching the legal-label decoder. Each image thus receives a label vector, with 60 dimensions for NICO++ object questions and six for context questions.

\paragraph{Retrieval and prediction.}
Prob.\ compares softmax probability vectors and Log.\ compares mean-centered log-probability vectors, both by cosine similarity. The four most similar bank entries enter the prompt with the nearest last. The four-shot prompt and legal-label decoder are identical to Raw. On NICO++, belief retrieval searches only bank entries of the query's question type and requires four distinct images.

\paragraph{Scoring checks.}
Packed label scoring agrees with separate full forward passes per label, with absolute errors below 0.1 for plausible labels ($\log p>-10$) and relative errors below 3\% for very unlikely labels. Batched scoring with 16 inputs agrees with single-input scoring on 16 of 16 test rows for DTD, CUB, Dogs, Pets and NICO++, with maximum probability differences of at most 0.036. On Aircraft, near ties between fine-grained labels let bf16 padding effects flip 2 of 8 test predictions, so Aircraft is scored one input at a time without padding, as in the Raw protocol.

\paragraph{Results.}
Tables~\ref{tab:supp_belief_all} through~\ref{tab:supp_belief_overlap} use 2000 paired bootstrap resamples with seed 73, resampling images with both questions on NICO++. Overall accuracy and Raw-minus-Log.\ intervals are reported in the main paper. Table~\ref{tab:supp_belief_all} retains only the additional Raw-minus-Prob.\ intervals. Prob.\ falls behind Raw mainly on initially wrong queries, where its neighbors follow the model's mistaken label. Log.\ preserves the relative ranking of labels and comes closer, yet it still trails on DTD, Dogs and NICO++, and its neighbors overlap with Raw's for only 11.0\% to 50.8\% of positions.

\begin{table}[H]
\centering\small
\caption{TaskVICL-Raw minus probability-vector belief retrieval (Prob.) on Qwen, with paired 95\% intervals. Absolute accuracies and the Raw-minus-Log. comparison are in the main paper.}
\label{tab:supp_belief_all}
\setlength{\tabcolsep}{5pt}
\begin{tabular*}{\linewidth}{@{\extracolsep{\fill}}lc@{}}
\toprule
Dataset & Raw $-$ Prob.\\
\midrule
DTD & $+8.03$ $[6.38,9.73]$\\
Aircraft & $+14.25$ $[12.48,16.02]$\\
CUB & $+17.29$ $[16.00,18.49]$\\
Dogs & $+9.56$ $[8.66,10.42]$\\
Pets & $+2.97$ $[2.21,3.79]$\\
NICO++ object & $+1.02$ $[0.19,1.94]$\\
NICO++ context & $+12.04$ $[9.35,14.54]$\\
NICO++ mean & $+6.53$ $[5.09,7.87]$\\
\bottomrule
\end{tabular*}
\end{table}

Table~\ref{tab:supp_belief_groups} separates the queries that the model answers correctly and incorrectly without demonstrations.

\begin{table*}[!tbp]
\centering\small
\caption{Raw minus belief retrieval on queries that Qwen classifies correctly or incorrectly without demonstrations, with paired 95\% intervals.}
\label{tab:supp_belief_groups}
\setlength{\tabcolsep}{4.5pt}
\begin{tabular*}{\textwidth}{@{\extracolsep{\fill}}lrllrll@{}}
\toprule
 & \multicolumn{3}{c}{Initially correct} & \multicolumn{3}{c}{Initially wrong}\\
\cmidrule(lr){2-4}\cmidrule(l){5-7}
Dataset & $n$ & Raw $-$ Prob. & Raw $-$ Log. & $n$ & Raw $-$ Prob. & Raw $-$ Log.\\
\midrule
DTD & 1261 & $+1.27$ $[0.00,2.62]$ & $+1.11$ $[0.08,2.14]$ & 619 & $+21.81$ $[17.77,26.01]$ & $+4.85$ $[1.78,7.92]$\\
Aircraft & 1077 & $+1.86$ $[-0.47,4.27]$ & $-0.84$ $[-2.88,1.11]$ & 2256 & $+20.17$ $[17.86,22.52]$ & $+0.40$ $[-1.33,2.13]$\\
CUB & 2154 & $+4.83$ $[3.39,6.27]$ & $+0.70$ $[-0.37,1.72]$ & 3640 & $+24.67$ $[22.97,26.37]$ & $+0.58$ $[-0.52,1.79]$\\
Dogs & 4667 & $+1.97$ $[1.14,2.76]$ & $+0.13$ $[-0.51,0.77]$ & 3913 & $+18.60$ $[16.97,20.16]$ & $+2.48$ $[1.30,3.68]$\\
Pets & 3026 & $+0.17$ $[-0.33,0.66]$ & $0.00$ $[-0.40,0.36]$ & 643 & $+16.17$ $[12.29,19.60]$ & $+2.64$ $[0.16,5.13]$\\
NICO++ object & 989 & $+0.10$ $[-0.77,0.98]$ & $+1.92$ $[1.11,2.83]$ & 91 & $+10.99$ $[3.30,19.78]$ & $+3.30$ $[-2.20,8.79]$\\
NICO++ context & 420 & $+0.48$ $[-2.51,3.46]$ & $+2.38$ $[-0.48,5.48]$ & 660 & $+19.39$ $[15.61,23.18]$ & $+14.70$ $[11.21,18.18]$\\
\bottomrule
\end{tabular*}
\end{table*}

Table~\ref{tab:supp_belief_overlap} reports neighbor overlap with Raw and the agreement between belief-based top labels and zero-shot predictions.

\begin{table}[H]
\centering\small
\caption{Top-four neighbor overlap with Raw, the share of queries with no shared neighbor, and the share of images whose top belief label equals the zero-shot prediction. Values are percentages.}
\label{tab:supp_belief_overlap}
\setlength{\tabcolsep}{2pt}
\begin{tabular*}{\linewidth}{@{\extracolsep{\fill}}lrrrrr@{}}
\toprule
 & \multicolumn{2}{c}{Prob.} & \multicolumn{2}{c}{Log.} & \\
\cmidrule(lr){2-3}\cmidrule(lr){4-5}
Dataset & Overlap & None & Overlap & None & Top $=$ ZS\\
\midrule
DTD & 18.0 & 47.6 & 42.4 & 13.6 & 98.1\\
Aircraft & 12.1 & 63.8 & 37.1 & 17.4 & 94.8\\
CUB & 20.4 & 44.2 & 50.8 & 5.8 & 96.9\\
Dogs & 13.6 & 57.4 & 45.7 & 9.2 & 96.3\\
Pets & 13.8 & 57.0 & 33.3 & 21.5 & 99.1\\
NICO++ object & 11.4 & 64.2 & 33.6 & 24.5 & --\\
NICO++ context & 3.5 & 88.0 & 11.0 & 65.7 & --\\
\bottomrule
\end{tabular*}
\end{table}

\begin{figure*}[t]
\centering
\includegraphics[width=\textwidth]{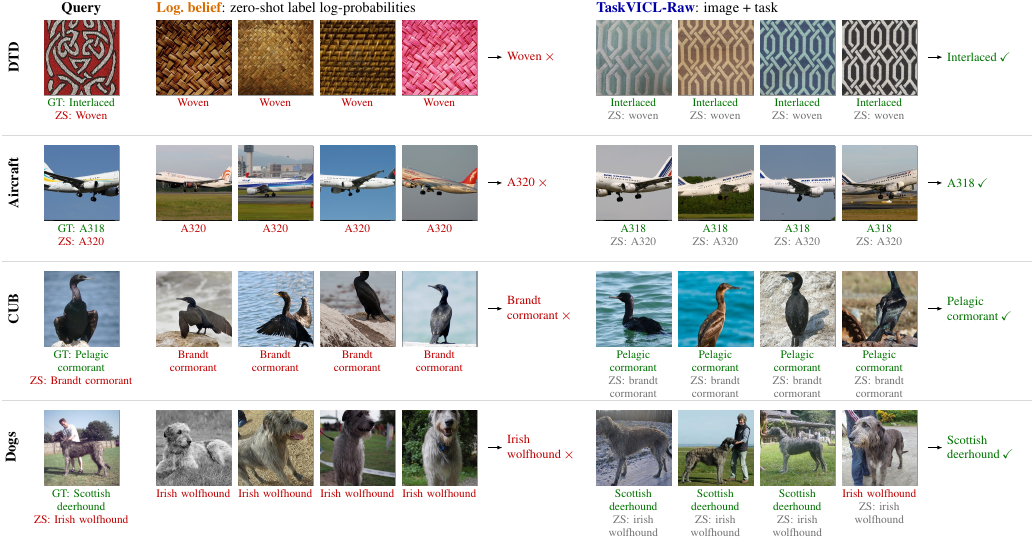}
\caption{Log.\ belief retrieval and TaskVICL-Raw on Qwen3-VL-4B. Log.\ ranks bank examples by the cosine similarity of centered zero-shot label log-probabilities. Notation follows Fig.~\ref{fig:supp_cls_rices}. Examples are selected among queries where zero-shot and Log.\ are wrong and Raw is correct.}
\label{fig:supp_cls_belief}
\end{figure*}

Figure~\ref{fig:supp_cls_belief} contrasts the Log.\ belief retriever with Raw on queries that both the model and Log.\ get wrong. Log.\ neighbors share the model's wrong belief and carry that wrong class as their reference answer, so the VLM repeats its zero-shot error. Raw's neighbors receive the same wrong zero-shot prediction but belong to the true class.

\end{document}